\documentclass{article} % For LaTeX2e
\usepackage{collas2023_conference,times}
\usepackage{easyReview}
\usepackage{booktabs}
\usepackage{caption}

\usepackage{amsmath,amsfonts,bm}

\def\eqref#1{equation~\ref{#1}}
\def\1{\bm{1}}

\DeclareMathAlphabet{\mathsfit}{\encodingdefault}{\sfdefault}{m}{sl}
\SetMathAlphabet{\mathsfit}{bold}{\encodingdefault}{\sfdefault}{bx}{n}

\newcommand{\figmotiv}[3]{
\begin{figure}[tb!]
    \centering
    \begin{tabular}{ll}
        \includegraphics[trim=#1, clip, width=#2\linewidth]{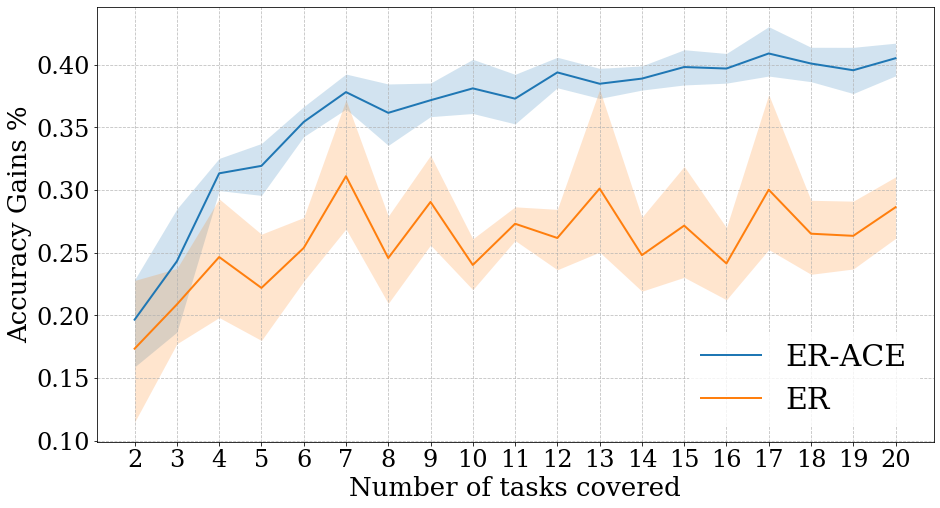} & \includegraphics[trim=#1, clip, width=#2\linewidth]{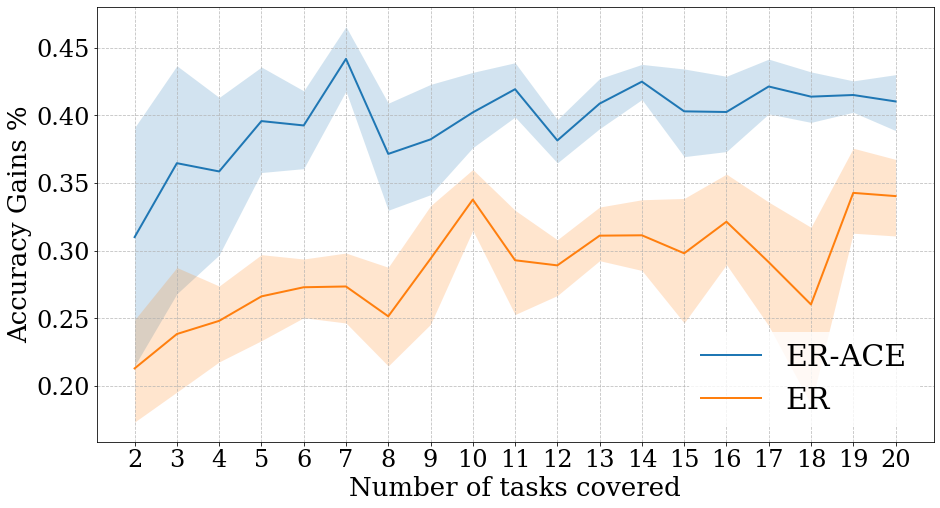} \\
    \end{tabular}
    \caption{#3}
    \label{fig:motivation}
\end{figure}
}

\newcommand{\figema}[3]{
\begin{figure}[tb!]
\centering
\begin{tabular}{cc}
    \includegraphics[trim=#1, clip, width=#2\linewidth]{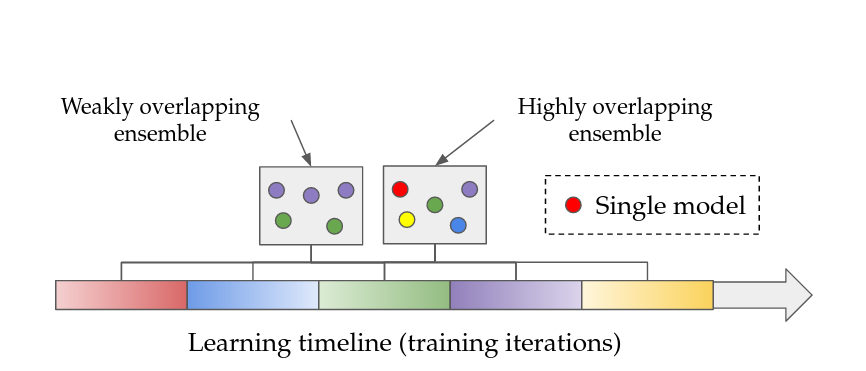} & \includegraphics[trim=#1, clip, width=#2\linewidth]{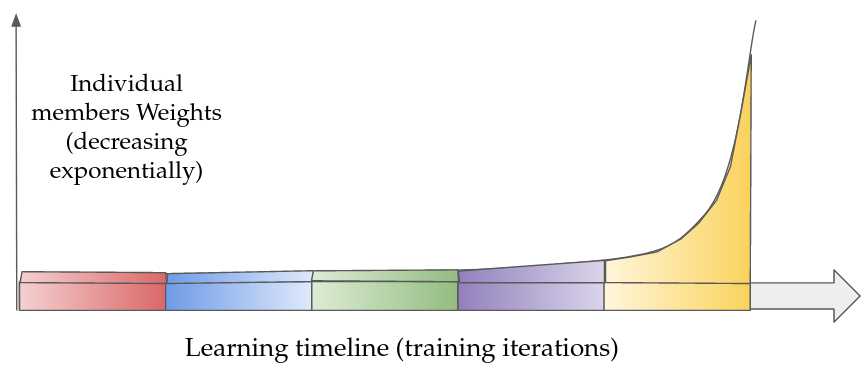}
\end{tabular}
\caption{#3}
\label{fig:ema_explained}
\end{figure}
}

\newcommand{\figstabace}[3]{
\begin{figure}[tb!]
\centering
\includegraphics[trim=#1, clip, width=#2\linewidth]{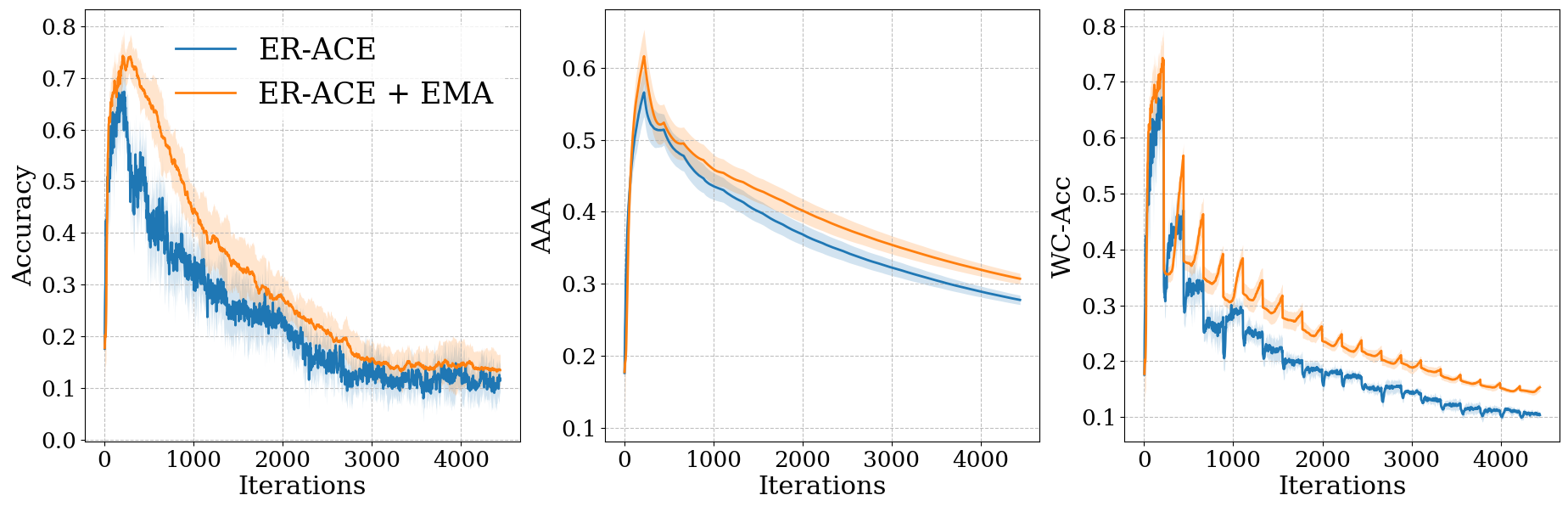}
\caption{#3}
\label{fig:stab_ace}
\end{figure}
}

\newcommand{\figstabrar}[3]{
\begin{figure}[tb!]
\centering
\includegraphics[trim=#1, clip, width=#2\linewidth]{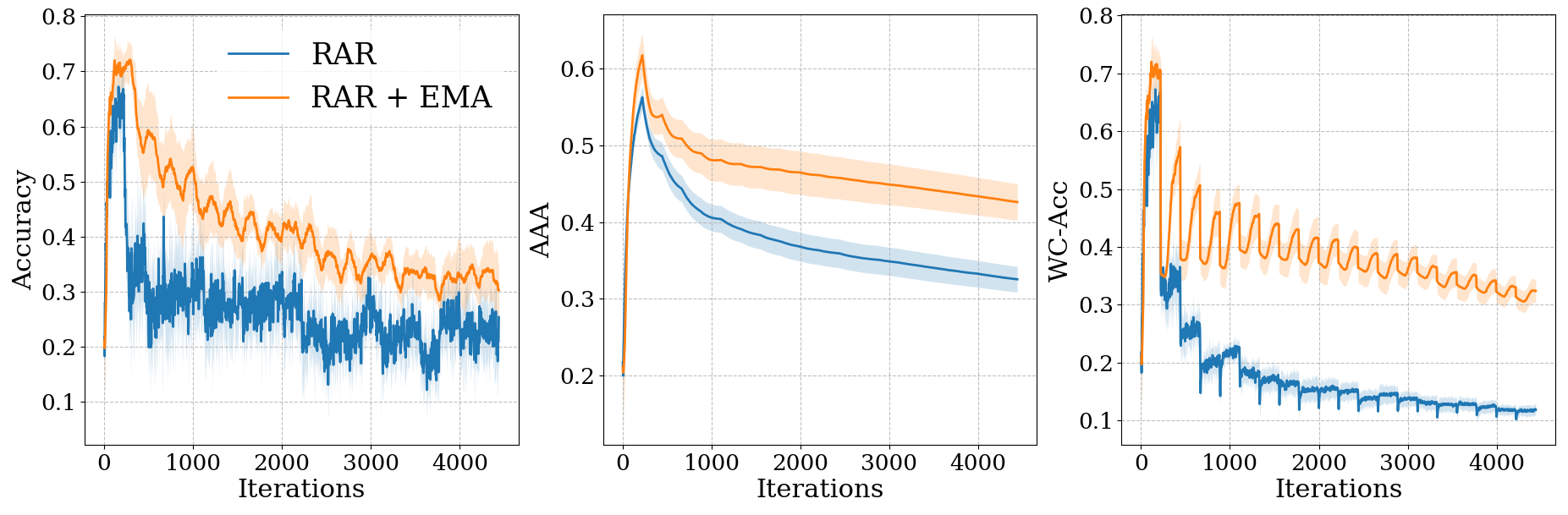}
\caption{#3}
\label{fig:stab_rar}
\end{figure}
}

\newcommand{\figimnet}[3]{
\begin{figure}[tb!]
\centering
\includegraphics[trim=#1, clip, width=#2\linewidth]{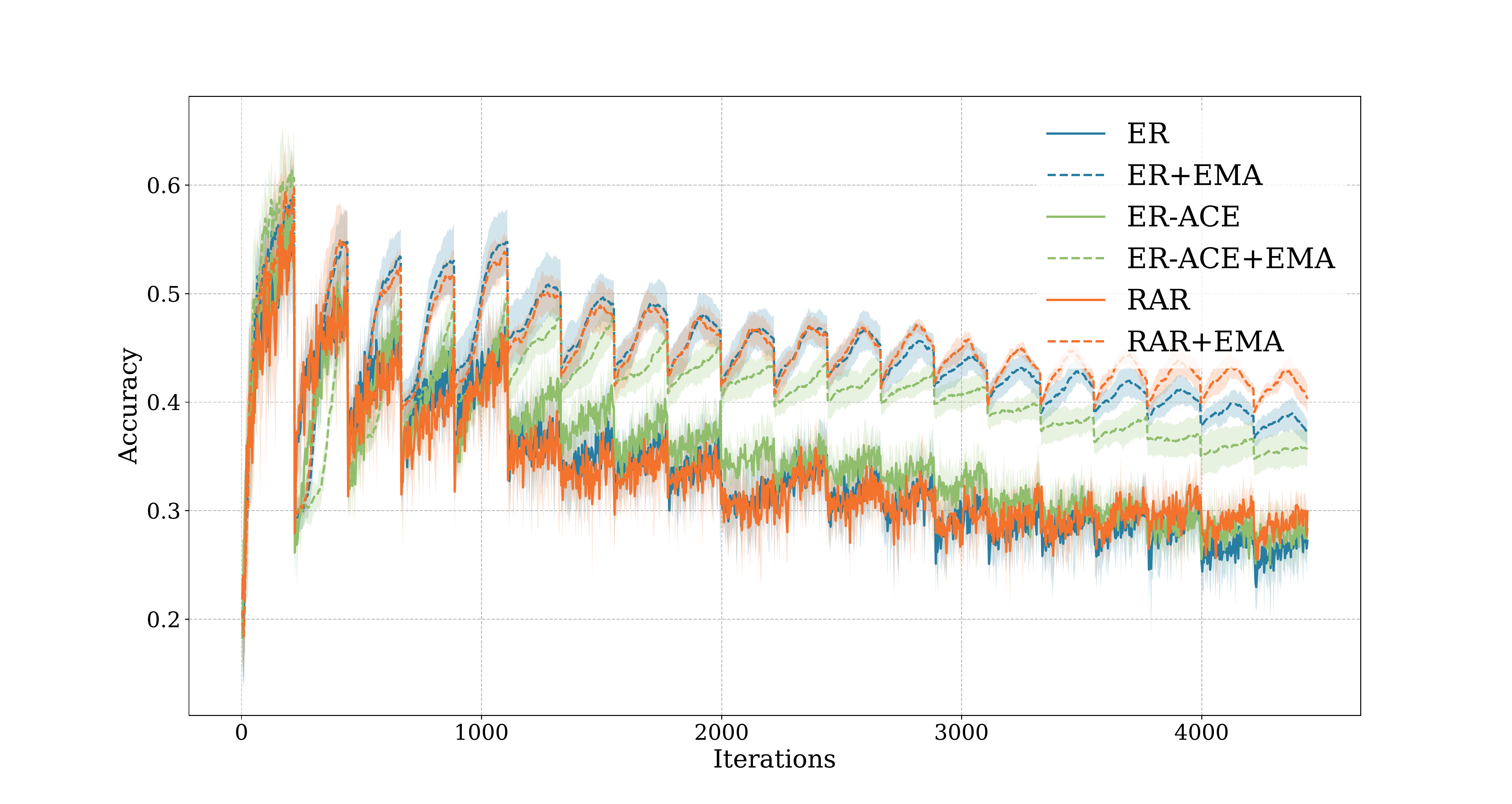}
\caption{#3}
\label{fig:comparison_imnet}
\end{figure}
}

\newcommand{\figcifarcien}[3]{
\begin{figure}[tb!]
\centering
\includegraphics[trim=#1, clip, width=#2\linewidth]{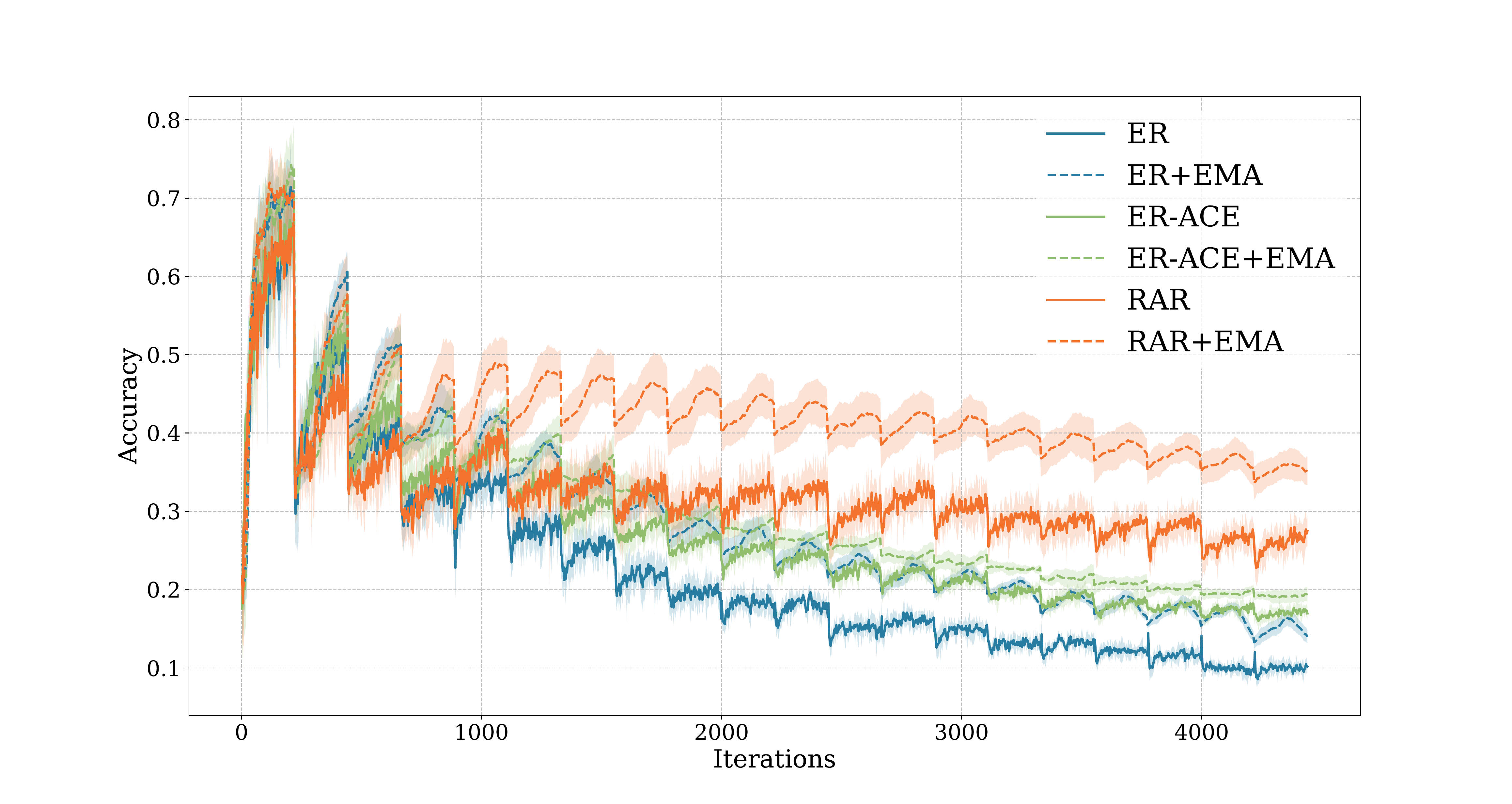}
\caption{#3}
\label{fig:comparison_cifar100}
\end{figure}
}

\newcommand{\figlambdas}[3]{
\begin{figure}[tb!]
\centering
\includegraphics[trim=#1, clip, width=#2\linewidth]{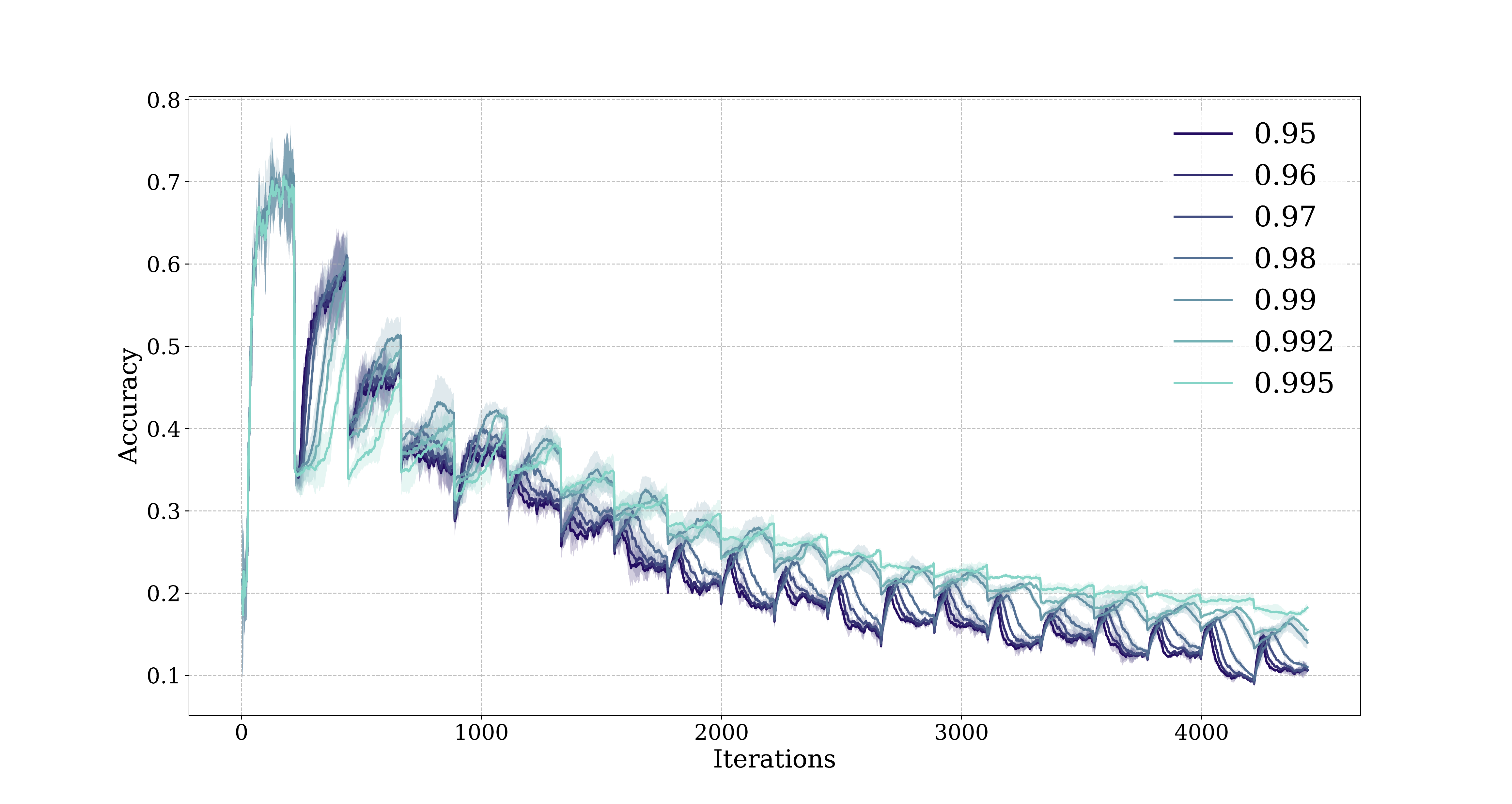}
\caption{#3}
\label{fig:comparison_lambdas}
\end{figure}
}

\newcommand{\tableensemblecomp}[1]{
\begin{table}[tb!]
\centering
\begin{tabular}{crr}
    \toprule
    Method & Accuracy & Memory (Mb) \\
    \midrule
    ER & 26.2 & 4 + 211  \\
    \midrule
    ER Naive Ensemble & 32.1 & 1840 + 211 \\
    \midrule
    ER + EMA & 36.3 & 8 + 211 \\
    \bottomrule
\end{tabular}
\caption{#1}
\label{tab:comp_ensemble}
\end{table}
}

\newcommand{\tableaccuracycifar}[1]{
\begin{table}[tb!]
\centering

\resizebox{\textwidth}{!}{%
\begin{tabular}{l|cccc|cccc}
\hline
& \multicolumn{4}{c|}{Split-Cifar10} & \multicolumn{4}{c}{Split-Cifar100} \\
Method & Acc $\uparrow$& $AAA^{val}$ $\uparrow$ & $\operatorname{WC-Acc^{val}}$$\uparrow$ & $RAG^{val}$$\downarrow$ & Acc $\uparrow$ & $AAA^{val}$$\uparrow$ & $\operatorname{WC-Acc}^{val}$$\uparrow$ & $RAG^{val}$$\downarrow$ \\
\hline
$i.i.d$ & 65.2 $\pm$ 1.8 & & & & 23.4 $\pm$ 0.7 &  \\
$i.i.d^{+EMA}$ & 67.3 $\pm$ 1.3 & & & & 25.8 $\pm$ 0.6 & \\
 & +2.1 & & & & +2.4
\\
\hline
$i.i.d_{w/tr}$ & 71.2 $\pm$ 1.4 & & & & 32.3 $\pm$ 1.0 & &\\
$i.i.d_{w/tr}^{+EMA}$ & 75.5 $\pm$ 1.3 & & & & 37.2 $\pm$ 0.8 & &\\
& +4.3 & & & & +4.9 & \\
\hline \hline
$ER$ & 37.5 $\pm$ 1.6 & 57.3 $\pm$ 0.6 & 26.0 $\pm$ 1.0 & 31.0 $\pm$ 5.6 & 9.9 $\pm$ 0.6 & 22.5 $\pm$ 0.8 & 5.8 $\pm$ 0.4 & 43.0 $\pm$ 3.4 \\
$ER^{+EMA}$ & 38.8 $\pm$ 1.3 & 59.9 $\pm$ 0.7 & 35.1 $\pm$ 1.1 & 9.9 $\pm$ 1.5 & 14.0 $\pm$ 0.5 & 29.2 $\pm$ 0.9 & 13.1 $\pm$ 0.7 & 5.6 $\pm$ 1.6 \\
& +1.2 & +2.6 & +9.1 & -21.1 & +4.1 & +6.7 & +7.3 & -37.4 \\
\hline
$MIR$ & 40.2 $\pm$ 2.8 & 54.0 $\pm$ 0.4 & 17.0 $\pm$ 0.6 & 57.1 $\pm$ 4.4 & 10.6 $\pm$ 0.7 & 22.8 $\pm$ 0.7 & 6.3 $\pm$ 0.4 & 39.6 $\pm$ 3.0 \\
$MIR^{+EMA}$ & 42.7 $\pm$ 2.1 & 55.9 $\pm$ 4.2 & 34.5 $\pm$ 1.5 & 19.7 $\pm$ 4.8 & 14.9 $\pm$ 0.4 & 28.8 $\pm$ 0.9 & 14.3 $\pm$ 0.6 & 5.4 $\pm$ 1.4\\
& +2.5 & +1.9 & +7.5 & -37.4 & +4.3 & +6.0 & +8.0 & -34.2 \\
\hline
$\operatorname{\mathit{ER-ACE}}$ & 50.2 $\pm$ 1.2 & 62.7 $\pm$ 1.4 & 34.6 $\pm$ 2.0 & 29.9 $\pm$ 4.8 & 16.5 $\pm$ 0.7 & 27.7 $\pm$ 0.6 & 10.3 $\pm$ 0.4 & 38.7 $\pm$ 3.3 \\
$\operatorname{\mathit{ER-ACE^{+EMA}}}$ & 51.5 $\pm$ 1.6 & 64.6 $\pm$ 1.4 & 48.7 $\pm$ 1.9 & 4.6 $\pm$ 0.8 & 19.0 $\pm$ 0.3 & 30.7 $\pm$ 0.6 & 15.3 $\pm$ 0.6 & 20.9 $\pm$ 3.5 \\ 
& +1.3 & +1.9 & +14.1 & -25.2 & +2.5 & +3.0 & +5.0 & -17.8 \\
\hline
$DER_{++}$ & 51.1 $\pm$ 3.0 & 56.4 $\pm$ 1.6 & 17.9 $\pm$ 0.6 & 64.5 $\pm$ 2.9 & 18.2 $\pm$ 0.7 & 24.7 $\pm$ 1.0 & 4.5 $\pm$ 0.6 & 75.8 $\pm$ 3.2 \\
$DER_{++}^{+EMA}$ & 53.1 $\pm$ 3.7 & 59.5 $\pm$ 1.7 & 36.4 $\pm$ 1.1 & 30.6 $\pm$ 3.5 & 23.2 $\pm$ 1.2 & 35.0 $\pm$ 1.3 & 19.7 $\pm$ 0.9 & 16.7 $\pm$ 2.3 \\
& +2.0 & +3.2 & +18.5 & -33.9 & +5 & +10.3 & +15.2 & -59.1\\
\hline
$RAR$ & 63.0 $\pm$ 1.4 & 67.3 $\pm$ 0.9 & 28.2 $\pm$ 2.9 & 55.1 $\pm$ 4.1 & 27.6 $\pm$ 1.3 & 32.5 $\pm$ 1.5 & 11.8 $\pm$ 0.9 & 57.0 $\pm$ 1.8 \\
$RAR^{+EMA}$ & \textbf{67.3 $\pm$ 0.8} & \textbf{72.4 $\pm$ 0.8} & \textbf{60.5 $\pm$ 1.1} & 10.1 $\pm$ 1.5 & \textbf{35.4 $\pm$ 1.2} & \textbf{42.6 $\pm$ 2.1} & \textbf{32.4 $\pm$ 1.7} & 7.9 $\pm$ 0.9 \\
& +4.3 & +4.9 & +32.3 & -40.0 & +7.8 & +10.1 & +20.6 & -49.9 \\
\hline
\end{tabular}
}
\caption{#1}
\label{tab:accuracies_cifar100}
\end{table}
}

\newcommand{\tableaccuracyimnetalt}[4]{

\begin{figure*}[tb!]
\begin{minipage}[c]{0.5\textwidth}
    \centering
    \resizebox{1.0\textwidth}{!}{%
    \begin{tabular}{l|cccc}\hline
    & \multicolumn{4}{c}{Split-MiniImagenet} \\
    Method & Acc $\uparrow$ & $AAA^{val}$$\uparrow$ & $\operatorname{WC-Acc}^{val}$$\uparrow$ & $RAG^{val}$$\downarrow$ \\
    \hline
    $i.i.d$ & 29.9 $\pm$ 2.2 &  & \\
    $i.i.d^{+EMA}$ & 34.9 $\pm$ 1.6 & \\
    & +5.0 &  \\
    \hline
    $i.i.d_{w/tr}$ & 32.3 $\pm$ 1.7 & \\
    $i.i.d_{w/tr}^{+EMA}$ & 39.9 $\pm$ 1.1 &  & \\
    & +7.6 &  \\
    \hline \hline
    $ER$ & 26.2 $\pm$ 0.2 & 33.9 $\pm$ 0.7 & 11.0 $\pm$ 0.9 & 59.6 $\pm$ 2.2  \\
    $ER^{+EMA}$ & 36.3 $\pm$ 1.1 & 44.3 $\pm$ 0.9 & 34.2 $\pm$ 0.6 & 7.9 $\pm$ 1.4 \\
    & +10.1 & +10.4 & +13.2 & -51 \\
    \hline
    $MIR$ & 27.3 $\pm$ 1.7 & 33.9 $\pm$ 0.4 & 9.6 $\pm$ 0.9 & 66.3 $\pm$ 2.3  \\
    $MIR^{+EMA}$ & 36.1 $\pm$ 1.2 & 43.5 $\pm$ 0.8 & 34.0 $\pm$ 1.4 & 8.6 $\pm$ 2.0  \\
    & +8.8 & +9.6 & +24.4 & -57.7 \\
    \hline
    $\operatorname{\mathit{ER-ACE}}$ & 27.4 $\pm$ 1.7 & 35.3 $\pm$ 0.5 & 12.9 $\pm$ 0.9 & 54.2 $\pm$ 2.6 \\
    $\operatorname{\mathit{ER-ACE^{+EMA}}}$ & 34.5 $\pm$ 0.8 & 41.0 $\pm$ 0.8 & 19.8 $\pm$ 1.1 & 44.6 $\pm$ 1.8 \\
    & +7.0 & +5.7 & +6.9 & -9.6 \\
    \hline
    $DER_{++}$ & 18.4 $\pm$ 1.8 & 20.3 $\pm$ 2.6 & 4.0 $\pm$ 0.4 & 78.5 $\pm$ 1.5  \\
    $DER_{++}^{+EMA}$ & 23.3 $\pm$ 2.3 & 29.4 $\pm$ 2.9 & 17.8 $\pm$ 3.3 & 25.0 $\pm$ 6.9 \\
    & +4.9 & +9.1 & +13.8 & -53.5 \\
    \hline
    $RAR$ & 29.1 $\pm$ 0.8 & 33.8 $\pm$ 0.8 & 11.4 $\pm$ 0.5 & 61.6 $\pm$ 2.1 \\
    $RAR^{+EMA}$ & \textbf{38.4 $\pm$ 0.8} & \textbf{44.9 $\pm$ 0.4} &  \textbf{35.6 $\pm$ 0.4} & 11.6 $\pm$ 1.7 \\
    & +9.3 & +11.1 & +24.2 & -60.0 \\
    \hline
    \end{tabular}
    }
    \captionof{table}{#1}
    \label{tab:accuracies_imnet}
\end{minipage}
\hspace{0.05\textwidth}
\begin{minipage}[c]{0.4\textwidth}
    \centering
    \begin{tabular}{c}
        \includegraphics[trim=#2, clip, width=#3\linewidth]{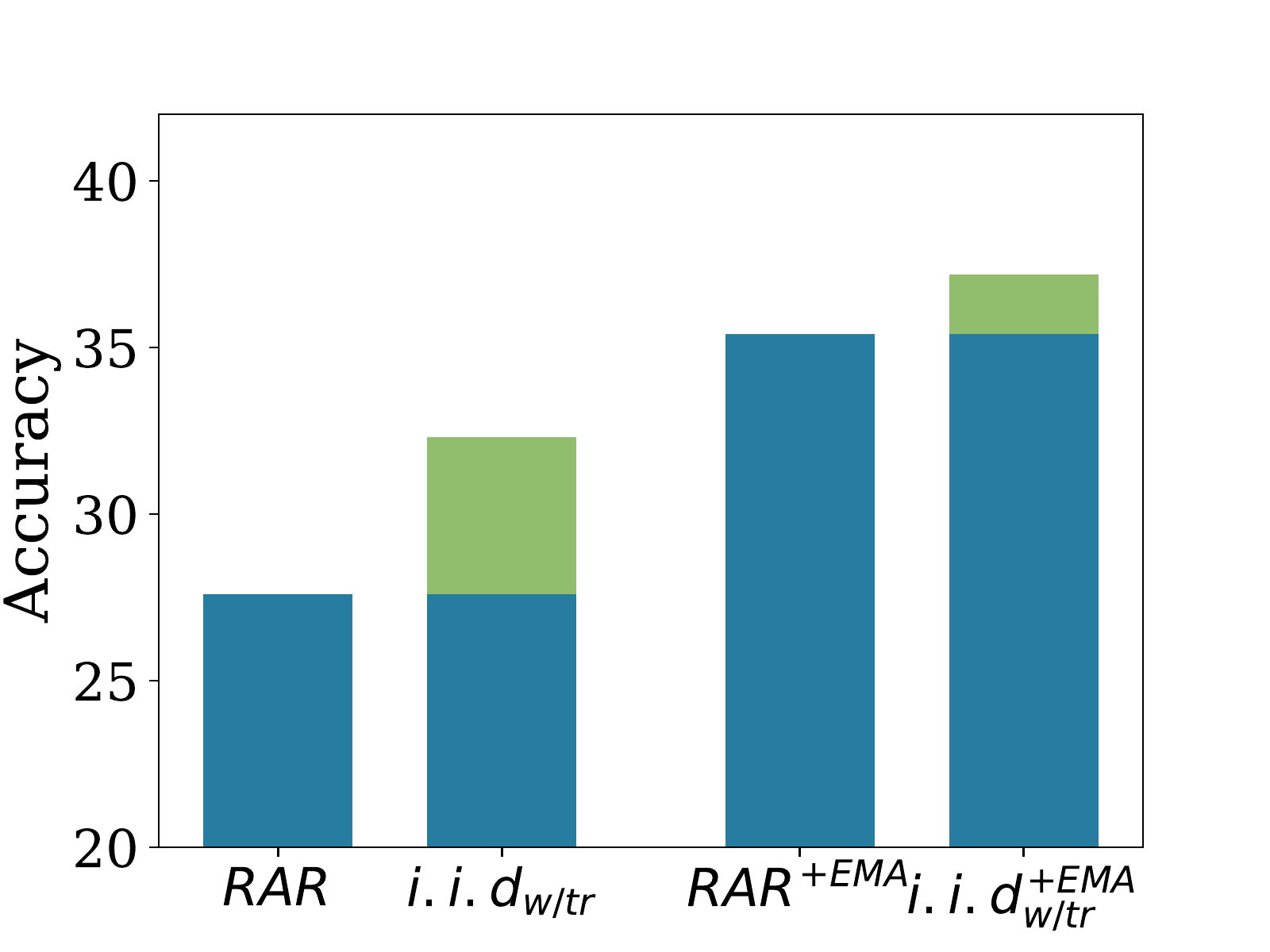} \\
        \includegraphics[trim=#2, clip, width=#3\linewidth]{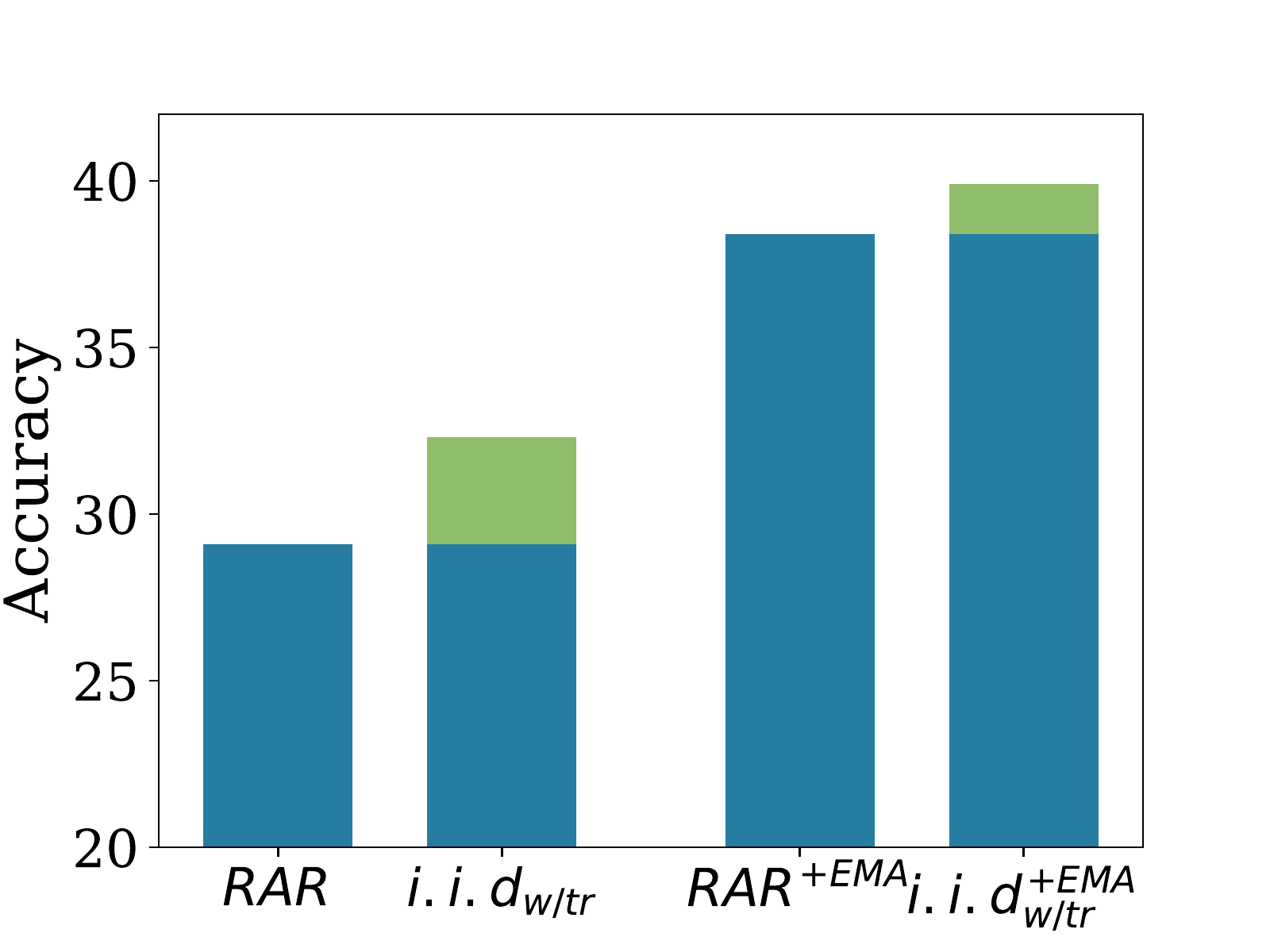} \\
    \end{tabular}
    \captionof{figure}{#4}
    \label{fig:barplots}
\end{minipage}
\end{figure*}
}

\newcommand{\figmatrix}[3]{
\begin{figure}[tb!]
\centering
\begin{tabular}{ccc}
    \includegraphics[trim=#1, clip, width=#2\linewidth]{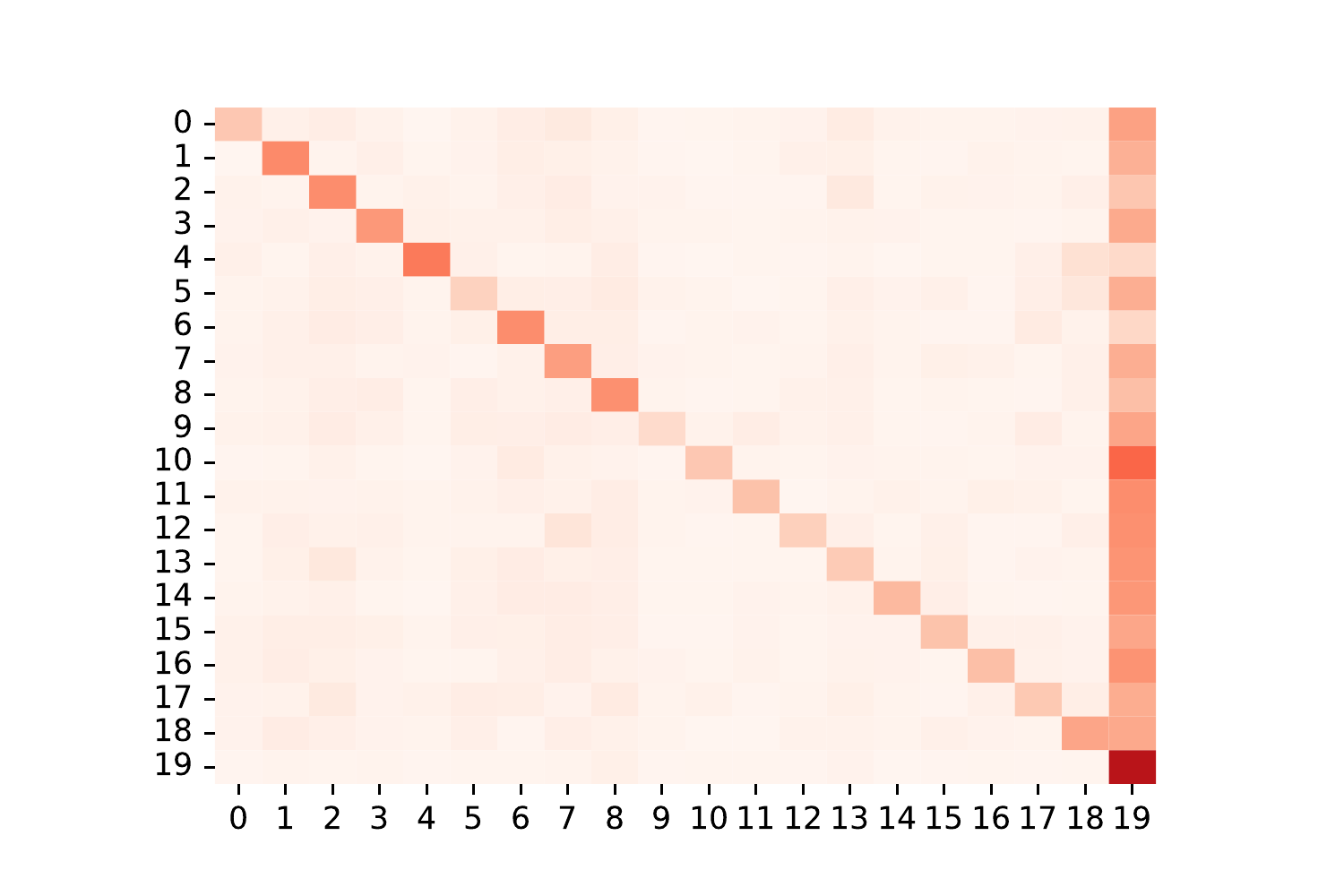} & \includegraphics[trim=#1, clip, width=#2\linewidth]{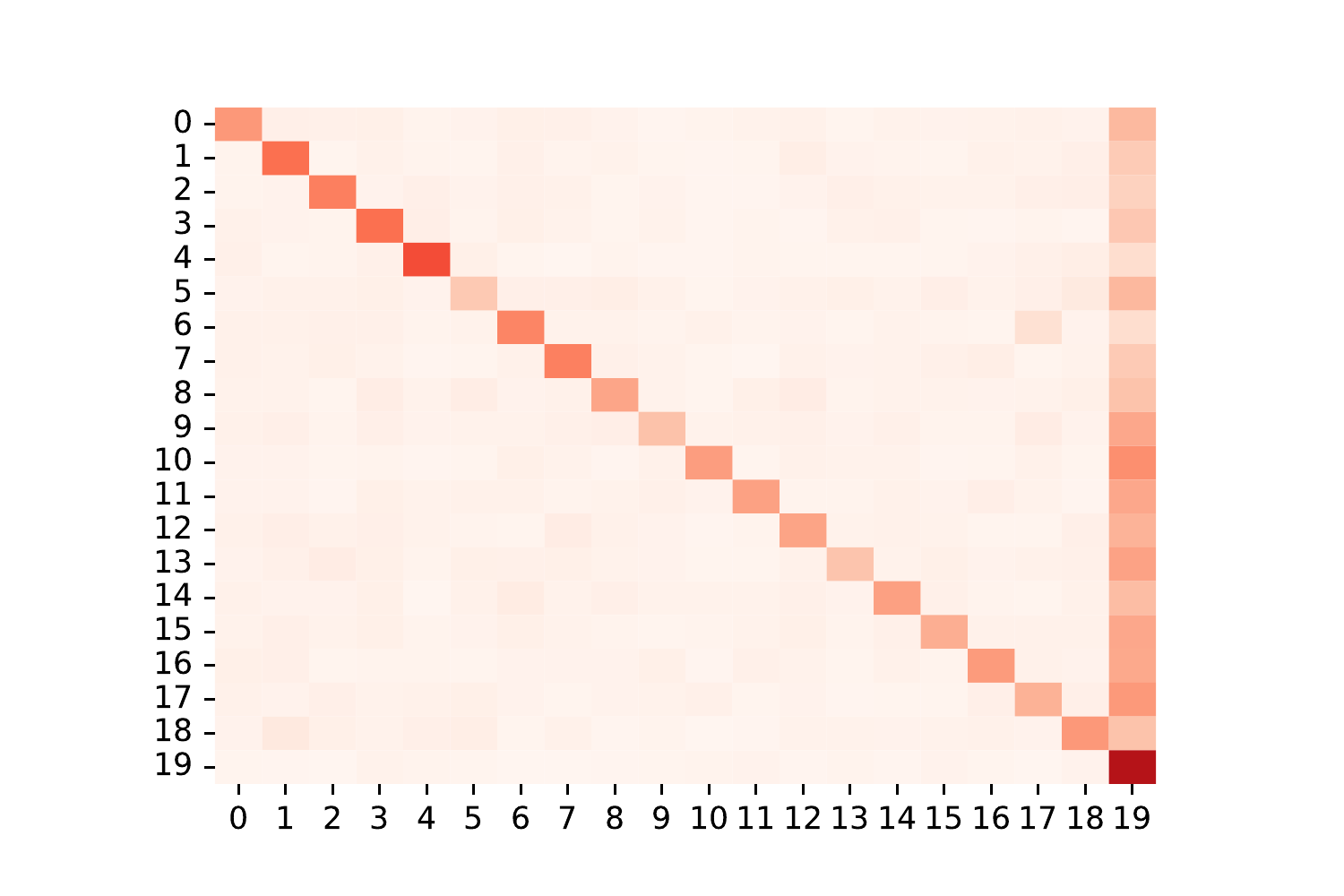} & \includegraphics[trim=#1, clip, width=#2\linewidth]{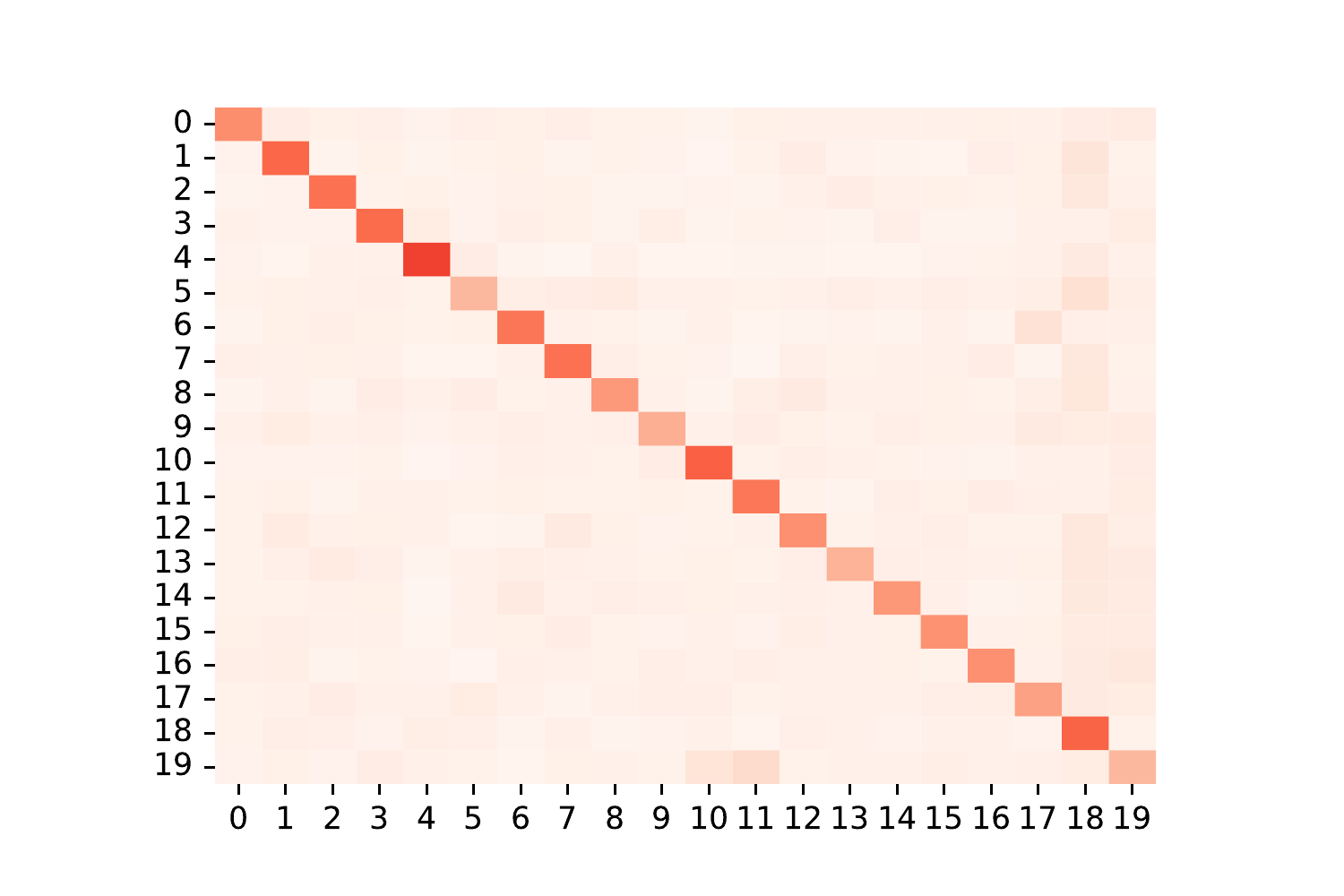}
\end{tabular}
\caption{#3}
\label{fig:confmatrix}
\end{figure}
}

\usepackage{hyperref}
\hypersetup{
    colorlinks=true,
    linkcolor=red,
    filecolor=magenta,
    urlcolor=blue,
    citecolor=purple,
    pdftitle={Overleaf Example},
    pdfpagemode=FullScreen,
}

\title{Improving Online Continual Learning Performance and Stability with Temporal Ensembles}

\author{Albin Soutif--Cormerais\\
Computer Vision Center\\
Universitat Autònoma de Barcelona\\
\texttt{albin@cvc.uab.cat}\\
\And % Use AND to have authors block one under the other
Antonio Carta \\
Department of Computer Science\\
University of Pisa\\
\texttt{antonio.carta@unipi.it}\\
\And % Use And to have authors side by side
Joost Van de Weijer \\
Computer Vision Center\\
Universitat Autònoma de Barcelona\\
\texttt{joost@cvc.uab.es}\\
}

\collasfinalcopy % Uncomment for camera-ready version, but NOT for submission.

\begin{document}

\maketitle

\begin{abstract}
Neural networks are very effective when trained on large datasets for a large number of iterations. However, when they are trained on non-stationary streams of data and in an online fashion, their performance is reduced (1) by the online setup, which limits the availability of data, (2) due to catastrophic forgetting because of the non-stationary nature of the data. Furthermore, several recent works \citep{caccia2022new,lange2023continual}  showed that replay methods used in continual learning suffer from the \textit{stability gap}, encountered when evaluating the model continually (rather than only on task boundaries). In this article, we study the effect of model ensembling as a way to improve performance and stability in online continual learning. We notice that naively ensembling models coming from a variety of training tasks increases the performance in online continual learning considerably. Starting from this observation, and drawing inspirations from semi-supervised learning ensembling methods, we use a lightweight temporal ensemble that computes the exponential moving average of the weights (EMA) at test time, and show that it can drastically increase the performance and stability when used in combination with several methods from the literature.

\end{abstract}

\section{Introduction}

\label{text:motivation}
Learning neural networks with backpropagation has been proven capable of good generalization properties even when using overparametrized networks \citep{overparametrized}. However, these good learning properties mainly occur when the data is provided in an independant and identically distributed manner. When learning on a stream which distribution varies over time, neural networks are known to suffer from \textit{catastrophic forgetting} \citep{mccloskey1989catastrophic,goodfellow2013empirical,kirkpatrick2017overcoming}, and tend to forget knowledge acquired in previous learning tasks. The field of \textit{continual learning} aims to address this problem. Generally, incremental learning separates the learning into distinct tasks (identified by a task-ID) that are encountered sequentially by the agent. A variety of settings have been introduced in continual learning in order to evaluate several aspects of the continual learning agent;  \textit{task-incremental learning}~\citep{til_survey,vandeven2019three}, and \textit{class-incremental learning}~\citep{cil_survey, belouadah2021comprehensive} are among the most popular. In this paper, we focus on the more challenging class-incremental setting, where the learner does not have access to the task-ID at inference time.

Model Ensembling, or the aggregation of predictions coming from different models, is a well studied and popular topic, both in the academic literature and in practical applications \citep{hansen1990neural, perrone1995networks, dietterich2000ensemble}. It is known to improve performance compared to using a single model. The success of ensembling methods has been found to depend on the functional diversity of the members and the efficiency of the resulting ensemble between them \citep{goodfellow2016deep, subspaces}.  In continual learning, the model learns the data task by task, being exposed to one task at a time. Therefore, models at different timesteps represent a functionally diverse set of models, each one locally adapted to the current task. Such functional diversity can be exploited by ensembling techniques. In Figure~\ref{fig:motivation},  we show ensembling results on two continual learning benchmarks. Here we apply a temporal ensemble of twenty models (chosen from a large number of models saved along the training trajectory). We can observe that ensembling leads to a significant performance improvement (over 40\% on both datasets), and that whenever we increase the number of tasks covered by the ensemble, gains improve. Therefore, in this article, we investigate the use of ensembles for continual learning and provide empirical results that show their benefit for continual learning settings. Importantly, we show that when using a practical and memory efficient ensembling method similar or even better results can be obtained.

Evaluation of continually learned agents typically occurs after learning a task. Recently, another way of evaluating has been proposed, coined \textit{continual evaluation}~\citep{caccia2022new,lange2023continual} or \textit{anytime inference}~\citep{koh2022online}. It aims to evaluate the agent's performance at any moment during learning. In this setting, \cite{lange2023continual} find that continual learning agents suffer from the \textit{stability gap}, where the performance on previous tasks decreases drastically at the start of learning a new task, before returning to normal when continuing training of the new task. This behavior is problematic in many real-world applications where the agent must be applied for inference while it is learning (i.e. for financial market forecasting, or online monitoring tasks). Ensembles can provide improved stability, as they can reduce the variance of the predictions and provide a more robust prediction. By combining multiple models, the errors of one model can be compensated for by the other models in the ensemble, leading to more stable performance.

Lastly, it is known that in class incremental learning, even when using replay, the network is prone to suffer from the task-recency bias, which is a prediction bias towards classes belonging to the last task. This phenomenon has been studied and tackled in many works \citep{belouadah2019il2m, wu2019large, hou2019learning}. Ensembling models biased towards different tasks has the potential to reduce the bias compared to the single models; we will analyse this in this paper. 

The reasons discussed in the introduction motivate us to investigate the potential of temporal ensembles in online continual learning. Specifically, we believe that they could offer improved stability, reduce task recency bias, and benefit from more functional diversity than in i.i.d learning. By exploring the performance of ensembles of models trained on sequential data, we aim to provide insights into the benefits and limitations of using such models in online continual learning settings. Our contributions are the following
\begin{itemize}
    \item We show that naively ensembling checkpoints from different continual learning tasks yields strong performance gains for online continual learning. In search for a practical ensembling method, we take inspiration from semi-supervised learning and apply a temporal ensembling method (i.e., an exponential moving average ensemble) as an evaluation model.
    \item We report the performance increase of the EMA ensemble in combination with several methods, showing a consistent increase in performance  (up to 9.3\% on Split-MiniImagenet). We also compute continual evaluation metrics and consistently notice a great increase in stability metrics resulting from the use of the EMA ensemble (up to 32.3\% on Split-Cifar10). We further observe a reduced task-recency bias of the EMA method.
\end{itemize}

\figmotiv{0 0 0 0}{0.475}{Relative accuracy gains (multiplicative in \%), compared to the worst performing ensemble, when naively ensembling 20 models coming from different learning tasks on Split-Cifar100 (Left) and Split-MiniImagenet (Right) for two online continual learning methods, classical replay ER and asymmetric cross entropy loss ER-ACE.
Results are reported as a function of \emph{number of covered tasks} which is defined as the number of tasks from which the ensembled models originate. The graph shows that diversity of the ensemble is important.}

\section{Related Work}

\subsection{Continual learning and online continual learning}

\label{text:related_work}

Popular continual learning scenarios often assume that data arrive in large batches of i.i.d data, with sharp distribution shifts happening whenever a new batch becomes available. We call this setting boundary-aware continual learning, due to the additional information provided by the arrival of a new task during training. Most of the continual learning literature focus on that setting, either by also providing the task-id at test time \citep{til_survey} (task-incremental learning), or not \citep{cil_survey} (class-incremental learning). We will focus on class-incremental learning in this paper.

Boundary-free online continual learning~\citep{aljundi2019task} removes this form of supervision by learning on a stream of small mini-batches. This means that the granularity of the data distribution can be refined. In practice, it is possible to keep tasks that define the granularity of the distribution, while letting the agent assume this distribution could change at any new mini-batch. This is the setting that we experiment on in this article. In MIR \citep{MIR}, the authors introduce a selection strategy for replay and select the samples that will be mostly penalised by the current update. In ER-ACE \citep{caccia2022new}, an asymetric cross-entropy loss is used on current data, using only the logits represented in the current mini-batch, while the classical cross-entropy loss is used on the replay buffer. \cite{rar} show that a strong baseline (called RAR) in online continual learning is repeatedly training on the available batch by sampling a new replay batch and applying data augmentations. In this article, we propose a simple improvement to these methods based on temporal ensembling which is applied at evaluation time.

Other, more classical class-incremental learning methods can be used in this setting, as long as they don't require knowledge of task-boundaries during training time. In ICaRL \citep{rebuffi2017icarl}, a selection strategy for storing the buffer sampled is used, along with a nearest mean classifier. In DER \citep{DER}, both the samples and the logits of these samples are stored and replayed using data augmentation. Other classes of approaches like EEIL \citep{castro2018end}, perform a balancing step at the end of training on each task, which makes them unsuitable to the boundary-free setting, unless the balancing is done before every evaluation session, in which case it could drastically increase the computation requirements. Other than these methods, we will focus on online continual learning methods in this paper.

\subsection{Ensembling in continual learning and temporal ensembling}

%Paragraph about ensembling
Aggregating the predictions coming from multiple trained models has been known for a long time as a process that results in increased performance compared to using the predictions of the individual models \citep{breiman1996bagging}. The group of trained models used for prediction is referred to as an ensemble of models. Such ensembles have been studied extensively in the literature \citep{hansen1990neural, perrone1995networks, dietterich2000ensemble}. Several challenges are associated with the creation of such ensembles, and a lot of the requirements of these ensembles seem at first incompatible with the constraints imposed by continual learning. In particular, naively creating an ensemble requires the training of several models that need to be stored in memory and trained independently, thus violating the memory and time constraints of continual learning. Nevertheless, some of these challenges have been already addressed in the literature. \cite{huang2017snapshot} relieve the constraint of having to train separate models by using checkpoints of the same training run and a cyclic learning rate schedule as the ensemble members. \cite{subspaces} train not only one network but a parametric subspace of networks which they can use to create an ensemble.

\cite{wen2020batchensemble} develop BatchEnsemble, a memory efficient way to create an ensemble of models by learning a shared weight matrix for all the members, and then a rank one matrix for each of the members. A member is then computed as the result of the hadamard product between the shared matrix and the rank one matrix. They later use this technique in task incremental learning, where they learn one member per task to be used at test time. \cite{single_model} lay the grounds of continual learning beyond the use of a single model. They study feasible ways of learning an ensemble of models continually, and compare a variety of ensembling techniques like BatchEnsemble \cite{wen2020batchensemble} or Subspace Learning \cite{subspaces}. They conclude that ensembling helps in the setting of task-incremental learning, and propose a method that make use of that property to increase the performance in that setting. Compared to our work, they focus more on the effect of adding more models into the ensemble but not so much on the effect of ensembling models coming from different tasks, and they operate in the task-incremental learning setting. \cite{lee2017overcoming} propose \textit{Incremental Moment Matching}, in which they compute the mean of the model weights in the weight space, in turn creating an approximate ensemble. In contrast to our work, they operate in the simpler task-incremental setting, in which task-ID is available at test time.

Temporal ensembling \citep{samuli2017temporal}, is a technique that consists in ensembling the predictions coming from different models on the training trajectory. In the original work, it was done by keeping an exponential moving average of the predictions of the model on the training data, but this technique was later refined in \citep{tarvainen2017mean}, where the authors chose to keep a running average of the weights instead of the predictions, and show that this leads to similar or even better performance, while relieving the constraint of having to update the running prediction for each datapoint at every iteration. In both of these works, the resulting ensemble prediction was used to improve the results in semi-supervised learning, where only a small fraction of the sample labels are available. This same Mean-teacher model has also been used successfully in several self-supervised learning works \citep{grill2020bootstrap, caron2021emerging}. In this article, we study the application of cheap temporal ensembles to the setting of online continual learning.

\section{Preliminaries}

\subsection{Continual Evaluation and The Stability Gap}

 In continual classification, a learning agent learns the parameters $\theta \in \Theta$ of a function $f: (\mathcal{X}, \Theta) \mapsto \mathcal{Y}$ from the image input space $\mathcal{X}$ to the label space $\mathcal{Y}$. It does so by observing a stream of data $\mathcal{S} = \{(x^1, y^1), (x^2, y^2), ... (x^n, y^n)\}$, where $x \in \mathcal{X}$ and $y \in \mathcal{Y}$. Each data tuple is drawn from a time varying distribution $(x^t, y^t) \sim \mathcal{D}_t$. In classical machine learning the training data distribution does not depend on time, but this is added as a constraint in continual learning. In both cases the goal of the agent is to perform well on new samples drawn from the joint distribution $\mathcal{D}$,  which is marginalized over past time. In practice, continual learning is simplified to allow for easier analysis by studying distributions that come from a discrete set and switch from one distribution to another (referred to as tasks $t \in \{t_1,...t_T\}$). In this paper, we will focus on class-incremental learning, where the learner does not have access to the task identifier $t$ at inference time.

 While all of the above simplifications make sense, they are still far from the human learning experience, and from fitting the requirements of many real-world applications. In comparison to the above, humans experience continuously time-varying distributions and continual evaluation. In order to address this, \cite{caccia2022new} and \cite{lange2023continual} lay the basis and encourage the study of continual evaluation of neural networks. In continual evaluation, the model is continuously evaluated during, instead of after each task. Interestingly, they noticed that the performance on previous tasks often drops at task shifts before coming back to a higher value later in training, this is what they refer to as the \emph{stability gap}.

\subsection{Continual Evaluation Metrics}%See where this is more relevant to put it

\label{text:metrics}

In this section, we present various metrics used in the online continual learning setting and that can help measure the stability and more generally, evaluate the performance of the agent over the course of its training. We denote $\mathbf{A}(E_i, f_t)$ the accuracy of $f_t$ (model at current iteration $t$), on the evaluation task $E_i$. The most common metric used in this scenario is the \textit{average anytime accuracy}, $AAA_t$ (See Eq.~\ref{eq:aaa}), used in many works \citep{osaka, caccia2022new, koh2022online}. While this metric does not focus on the worst-case performance, it is a nice indicator of the performance of the learning agent over the course of training. It measures the average accuracy on all tasks seen so far, and averages it over all training iterations. In \citep{lange2023continual}, a set of metrics is introduced to measure worst-case performance. These are particularly suited to assess the \emph{stability} of the algorithms. They first define the average minimum accuracy reached by previous tasks when learning task $T_k$, $\operatorname{min-ACC_{T_k}}$ (see Eq.~\ref{eq:min_acc}). It gives a good idea of the worst case performance of the agent on a given task. Then, the worst-case accuracy, $\operatorname{WC-ACC_{t}}$ (see Eq.~\ref{eq:wc_acc}), combines information from the minimum accuracy on previous tasks and the accuracy on the current task. $\operatorname{WC-ACC_{t}}$ summarizes the trade-off between stability (accuracy on previous task data) and plasticity (accuracy on current task data). This metric is upper-bounded by the average accuracy. Here, $t$ is the current iteration, $T_k$ the current task (at iteration t) and $t_{|T_{i}|}$ is the iteration at the end of learning $T_i$. Since $\operatorname{WC-ACC}_t$ is upper bounded by the average accuracy, we also report a new metric which is the relative gap between the latter and average accuracy $\operatorname{Acc}_t$ as defined in Equation \ref{eq:rel_wc}, we name it \textit{Relative Accuracy Gap} (RAG) since this measures the relative gap between worst-case accuracy and average accuracy. This metric can then be fairly compared across various methods that have different average accuracy.

\begin{equation}
    \operatorname{AAA_{t}} = \frac{1}{t}\sum_{j=1}^{t}\frac{1}{k}{\sum_{i=1}^{k}{\mathbf{A}(E_i, f_j)}}
    \label{eq:aaa}
\end{equation}

\begin{equation}
    \operatorname{min-ACC_{T_k}} = \frac{1}{k-1}\sum_{i}^{k-1}{\min\limits_{n}} \ \mathbf{A}(E_i, f_n), \forall n :   \; t_{|T_{i-1}|} < n \leq t
     \label{eq:min_acc}
\end{equation}

\begin{equation}
    \operatorname{WC-ACC_{t}} = \frac{1}{k}{A}(E_k, f_t) +  (1 - \frac{1}{k}) \operatorname{min-ACC_{T_k}}
    \label{eq:wc_acc}
\end{equation}

\begin{equation}
    \operatorname{Acc_t} = \frac{1}{k}\sum_{i}^{k}{A(E_i, f_{t})}
    \quad \mathrm{and} \quad
    \operatorname{RAG_t} = \frac{\operatorname{Acc_t} - \operatorname{WC-ACC_{t}}}{\operatorname{Acc_t}}
    \label{eq:rel_wc}
\end{equation}

\section{Method: Exponential Moving Average ensemble (EMA)}

\label{text:method}

\tableensemblecomp{Comparison on Split-MiniImagenet (20 tasks) of the naive ensembling of checkpoints taken along the training trajectory of a replay method every 10 iterations, against the use of EMA ensemble. For clarity, we divide the memory footprint into the one for the models and the one for the replay buffer (model + buffer).}

In the introduction (see Figure~\ref{fig:motivation}) we have shown that ensembles can greatly improve performance, however, they come at a significant increase in memory usage which is in direct conflict with the memory requirements typically imposed on continual learners. Indeed, both continual learning and online learning impose memory constraints since they do not allow retaining more than a fixed amount of data coming from the stream of data. Therefore, in this section, we look into methods to reduce the memory usage, while maintaining the advantages of model ensembling.

Several works have focused on reducing the memory constraint of ensembles \cite{wen2020batchensemble, subspaces, tarvainen2017mean}, some did so notably by averaging the models in weight space instead of aggregating the predictions in the functional space. While it is not clear under which condition such manipulation of the weights can form a model that performs similarly to the ensemble of the summed members, several works have shown practical working cases. It is possible to perform such a summation~\citep{subspaces} whenever two models are connected by a linear path of low loss \citep{frankle2020linear}. \cite{tarvainen2017mean} propose instead to do a weighted sum of an infinite amount of checkpoints by giving older checkpoints less important weight, decreasing exponentially with the distance.

For a potential use of these ensembles in continual learning we are interested in having an ensemble that covers many tasks (see Section~\ref{text:motivation}), and one that is cheap to store and compute. The solution adopted in \citep{tarvainen2017mean}, for semi-supervised learning, fits well to this task since it requires storing only one additional model and is able to ensemble models from all of the previous iterations. Consider a function $f(x) = f(x, \theta^{t})$ with learnable weights $\theta^{t}$ (at training iteration t), the exponential moving average (EMA) of it's weights is defined as:
\begin{equation}
    \theta^{t}_{ema} = \lambda\theta^{t-1}_{ema} + (1 - \lambda)\theta^{t},
    \label{eq:mean}
\end{equation}
where $\lambda$ is a user-defined hyperparameter comprised between 0 and 1, which sets the importance of the current model in the running average compared to the one of the previous models, $\theta^{t-1}_{ema}$ is the value of the moving average at the previous iteration, and $\theta^{t}$ is the weights of the training model at iteration $t$ ($\theta^{t}$ can be computed with existing online methods, such as ER \citep{tiny} or MIR \citep{MIR}. This implicit definition can also be written as an explicit sum over all the previous model weights:
\begin{equation}
    \theta^{t}_{ema} = \sum_{i=1}^{t}{(1-\lambda)\lambda^{t-i}\theta^{i}} + \lambda^{t}\theta^{0}_{ema}.
    \label{eq:explicit}
\end{equation}
This means that the ensemble formed by the sum virtually covers all the previously encountered tasks. However, exponentially less weights will be reserved to older tasks, potentially reducing the effective diversity of the ensemble. Nevertheless, the motivational experiment we conducted on Split-Cifar100 show that after some number of tasks covered by the ensemble, the accuracy gained by covering more and more tasks is less important (sublinear growth). So covering a small number of tasks with the ensemble can be sufficient to get satisfying performance gains. This motivated us to analyze the exponential moving average model in the setting of online continual learning. 

In Table~\ref{tab:comp_ensemble} we compare Naive Ensembling, discussed in Section~\ref{text:motivation}, with the EMA model when combined with Experience Replay~\cite{tiny}. As can be seen, the EMA model significantly reduces the memory usage (requiring only one additional model). Remarkably, ER+EMA outperforms the Naive Ensemble. This could be caused by the fact that ER+EMA combines many more models, and because of the non-linear weight assignment to the various models (see Appendix Figure~\ref{fig:ema_explained}), where EMA assigns more weight to the last (and better) models in the training trajectory. 

While the EMA approximate ensemble is commonly used in the literature and has been proven to give good performance \citep{tarvainen2017mean, grill2020bootstrap, caron2021emerging}. Other weight summing schemes could be tried to compute an approximate ensemble. Such schemes are not necessarily expected to work in classical offline learning since summing models that are far away from each other in the weight space is not guaranteed to work. However, since online learning only performs a few training iterations on the task at hand (compared to offline learning), we can expect the models to be closer from each other and thus allow other summing techniques to work. In order to explore the possibilities of such summing techniques in online continual learning, we compare several weighting techniques. Since we are working in the online learning setting and under the constraint of storing only one additional model, we choose to compute the approximate ensemble following Equation~\ref{eq:explicit_other},
\begin{equation}
    \theta^t_{ensemble} = \frac{1}{\sum_{i}^{t}{w_i}}\sum_{i}^{t}{w_i\theta^i},
    \label{eq:explicit_other}
\end{equation}
and use this ensemble instead of the EMA ensemble. This formulation allows for more freedom in the choice of the weighting scheme, but the formulation used by EMA can also be expressed under this form, we identify the equivalent weight $w_i$ for the EMA ensemble using Equation~\ref{eq:explicit} as being $w_i = \frac{w_{i-1}}{\lambda}$. We update at every iteration the weighted sum of the models and normalize it by the sum of the weights to avoid exploding model weights. In Appendix (Figure \ref{fig:weighting_schemes}) we provide a comparison of the weight distribution for the different strategies we tried.

\begin{table}[tb]
    \centering
    \begin{tabular}{|c|c|c|c|c|}
    \hline
    & $w_i$ & ER & ER-ACE & RAR \\
    - & - & 9.9 $\pm$ 0.6 & 16.5 $\pm$ 0.7 & 27.6 $\pm$ 1.3 \\
    EMA $\lambda$ = 0.99 & $\frac{w_{i-1}}{\lambda}$ & 14.0 $\pm$ 0.5 & 19.0 $\pm$ 0.3 & 35.4 $\pm$ 1.2 \\
    EMA $\lambda$ = 0.995 & $\frac{w_{i-1}}{\lambda}$ & 18.0 & \textbf{20.1} & \textbf{36.8}  \\
    EMA $\lambda$ = 0.999 & $\frac{w_{i-1}}{\lambda}$ & \textbf{18.3} & 18.8 & 31.2 \\
    Uniform & 1 & 13.1 & 12.7 & 17.3 \\
    Linear & i & 16.4 & 16.3 & 25.4 \\
    Logarithmic & $w_{i-1} + log(i)$ & 16.6 & 16.6 & 26.2 \\
    Quadratic & $w_{i-1} + i^2$ & 18.2 & 18.8 & 31.5 \\
    \hline
    \end{tabular}
    \caption{Comparison of different approximate ensembling methods inspired from the EMA approximate ensembling method. Each ensembling technique was tried on Split-Cifar100 in combination with ER \citep{tiny} ER-ACE \citep{caccia2022new}, and RAR \citep{rar}. The second column indicates how the weights for the current model are computed at each step. The first row indicates the results when no ensembling technique is used.}
    \label{tab:recursive}
\end{table}

In Table \ref{tab:recursive}, we present the results of combining an ensemble computed with each weighting scheme with three methods from the literature on the Split-Cifar100 dataset (See Section \ref{text:experiments} for explanations about the experimental settings). We see that the EMA model gets the best performance overall, especially with $\lambda = 0.995$\footnote{These parameters are not the one we use in the main results of Section \ref{text:results} (we use $\lambda = 0.99$). For a discussion on hyperparameter choices, refer to Section \ref{text:hp}}. However, we get surprisingly high results with linear, logarithmic, and quadratic weighting, especially in combination with ER, but linear and logarithmic weighting fail to give an advantage when combined with the more advanced methods ER-ACE and RAR. Quadratic weighting obtains the closest resuts to the EMA methods, we provide a more detailed comparison of Quadratic against EMA method in the Appendix (Figure.~\ref{fig:wc_acc_quad}). Uniform weighting is the scheme that performs worst across the board. which can be understood since it gives equal weight to the first models than to the last models whereas the last models have been trained for a longer time and thus are expected to have better performance.

\section{Experiments}

\label{text:experiments}

\textbf{Datasets.} We perform experiments on 3 datasets. \textbf{Cifar-10} is a 10-class dataset that contains 60000 images of size 32 by 32 and 3 color channels ~\citep{cifar}. \textbf{Cifar-100} has the same image dimensions and number of images but with 100 classes. \textbf{Mini-Imagenet} \citep{vinyals2016matching} is a 100-class version of \textbf{ImageNet} ~\citep{Imagenet}, that contains 60000 images which are rescaled to 84 by 84. We split these datasets into 5, 20 and 20 tasks respectively, each containing a mutually exclusive set of classes.

\textbf{Scenario.} We present results in the online class-incremental setting. When continual evaluation is performed, we evaluate after each unique mini-batch. All the compared methods are using a replay buffer, with a fixed memory size of 1000 exemplars for Cifar-10, 2000 exemplars for Cifar-100 and 10000 for Mini-Imagenet (as in ~\citep{caccia2022new}). Each training mini-batch is formed out of half of exemplars from previous tasks and half from the current task.

\textbf{Methods.} We compare the performance of five replay methods. ER-ACE \citep{caccia2022new}, MIR \citep{MIR}, RAR \citep{rar}, DER \citep{DER} are described in Section \ref{text:related_work}, while ER \citep{tiny} is the vanilla replay baseline. We display their performance along with the one of their EMA augmented version on the studied datasets. Additionally, we report the results of an $i.i.d$ reference method that is allowed the same memory and computational budget as the compared methods, but for which the data arrives in an independent and identically distributed manner (in contrast to the continual learning manner where data arrives split by split). Since some of the methods included in the comparison make use of input transformations (RAR), we also include the results of the $i.i.d_{w/tr}$ reference method, which uses the same input transformations.

\textbf{Training and Implementation Details.} For all datasets we use a slim version of Resnet-18 as done in \citep{gem} and perform 3 passes per mini-batch using Stochastic Gradient Descent with a learning rate of 0.1, and batch size of 32. We run each experiment for six seeds and report the mean and standard deviation. For DER, we stick to the parameters used in the original paper for CIFAR10 ($\alpha=0.1$ and $\beta=0.5$). For the EMA ensemble, we chose a momentum parameter of $\lambda=0.99$. More details on the choice of this parameter can be found in the Appendix Section~\ref{text:hp}. We make use of the Avalanche framework \citep{lomonaco2021avalanche} for all experiments. We make the code available at: \url{https://github.com/AlbinSou/online_ema}.

\textbf{Metrics.} For every method, we report the final average accuracy but also the continual evaluation metrics described in Section~\ref{text:metrics}, that we computed on a held out validation set after training on each new mini-batch. The validation dataset contains $5\%$ of the total training data. We report both $\operatorname{AAA_{T_{final}}}$ and $\operatorname{WC-ACC_{T_{final}}}$ in the tables, where $T_{final}$ is the last training iteration. We also report $\operatorname{WC-ACC_{t}}$ at every iteration in the figures. The final value of the $RAG$ metric that we define in Equation \ref{eq:rel_wc} is reported in percent. Note that we do not make use of this validation data to tune hyperparameters but just to compute the continual evaluation metrics.

\section{Results}

\label{text:results}

\figstabace{0 0 0 0}{0.9}{Validation accuracy on task 1 data (Left), Average Anytime Accuracy $AAA_t$ (Middle) and $\operatorname{WC-ACC}_t$ (Right) for ER-ACE and its EMA augmented version on Split-Cifar100, using 2000 memory. Mean and standard deviation are computed over 6 runs.}

\figstabrar{0 0 0 0}{0.9}{Split-Cifar100, validation accuracy on task 1 data (Left), Average Anytime Accuracy $AAA_t$ (Middle) and $\operatorname{WC-ACC}$ (Right), for RAR and its EMA augmented version, using 2000 memory. Mean and standard deviation are computed over 6 runs.}

\figimnet{10 10 10 10}{0.9}{Comparison of the average accuracy on the validation data of three methods trained on Split-MiniImagenet (20 tasks), and their EMA augmented version. EMA performance is indicated with the dotted lines. We report the mean and standard deviation over 6 runs.}

On \textbf{Split-Cifar10} the EMA ensemble offers consistent improvements across all methods (see Table~\ref{tab:accuracies_cifar100}), especially for RAR, where it offers a 4.3\% improvement in final average accuracy. We hypothesise that the gains of EMA are smaller than on Split-Cifar100 and Split-MiniImagenet because in that case the EMA ensemble weights cover less tasks than in the case of the other two datasets (5 tasks but the same number of training iterations than Split-Cifar100). Nevertheless, the stability metrics are greatly improved also for this dataset.

\tableaccuracycifar{Comparison of the final average accuracy on the test set and continual evaluation metrics on the validation set for various methods with their EMA model augmented version, on Split Cifar10 (Left) and Split Cifar100 (Right).}

On \textbf{Split-Cifar100} (See Table~\ref{tab:accuracies_cifar100}) the EMA ensemble offers considerable improvements for ER, RAR, DER and MIR (from 4.0-7.8\%). The improvements are less consequent for $\operatorname{ER-ACE}$, but still important (2.5\%). We hypotesize that the smaller gain is due to the smaller task-recency bias of $\operatorname{ER-ACE}$. This is illustrated in Figure~\ref{fig:stab_ace} and Figure~\ref{fig:stab_rar}. In these two figures on the left, we see that the performance of RAR on task 1 data drops instantly after learning task 1, which means it has been traded for accuracy on task 2, it suffers from the task-recency bias. Whereas for ER-ACE, the performance on task 1 does not drop as much after learning task 1, thus the gap with the EMA augmented version is not as important. This is also reflected in Table \ref{tab:accuracies_cifar100}. In these figures as well, we can see how the use of the EMA model improves the stability, both by looking at the $\operatorname{WC-Acc}$ but also at the reduced accuracy variations on a single task.

\tableaccuracyimnetalt{Comparison of the final average accuracy on the test set and continual evaluation metrics on the validation set for various methods along with their EMA model augmented version, on Split-MiniImagenet.}{0 0 0 0}{0.8}{Comparison of previous state-of-the-art method in online continual learning $RAR$ against the reference method $i.i.d_{w/tr}$ on Split-Cifar100 (Top) using 2000 memory and Split-Minimnet using 10000 memory (Bottom). The performance gap is indicated in green, and is greatly reduced by the use of EMA.}

On \textbf{Split-MiniImagenet}, we see the largest performance improvements. This time, RAR sees similar gains than ER (9.3\% and 10\%), while $\operatorname{ER-ACE}$ gains are also more important than in the case of Split-Cifar100. This is coherent with the observations that we had in the motivational experiments (Figure~\ref{fig:motivation}) that the gains from ensembling are slightly more important in the case of Split-MiniImagenet. This might also be due to the memory size used that is different from the one used for Split-Cifar100. In Figure~\ref{fig:comparison_imnet}, we show a comparison of three methods and their EMA augmented version. We notice that in that case, the use of ER-ACE hurts the performance of the EMA model, which performs worse than just using ER and EMA. Also, for ER and RAR, we notice various bumps in the validation accuracy \footnote{More generally we observe these bumps on all the studied datasets and report them for Split-MiniImagenet and Split-Cifar100 in the Appendix Figure~\ref{fig:comparison_imnet} and Figure~\ref{fig:comparison_lambdas}} of the EMA model that are not present in the current model accuracy. We believe these bumps occur when the previous task bias is compensated by bias towards the current task. The location and width of these bumps depend on the $\lambda$ parameter chosen for the exponential moving average (for more analysis see Appendix~\ref{fig:comparison_lambdas}).

\textbf{Effect on the task-recency bias:} In Figure \ref{fig:confmatrix}, we display the task confusion matrices of RAR, for the final training model, the final EMA model, and the EMA model taken at the tip of the bump (selected using a hold-out validation memory mechanism). The matrix shows the number of test images from a particular task (y-axis) that are classified as being from another task (x-axis). For the last training model, a lot of samples are predicted to be in the last task as indicated by the last column, which shows an important task-recency bias. For the final EMA model, the task-recency bias is also present though slightly diminished, but for the best selected EMA model, it is almost absent, confirming our hypothesis about the origin of the bump\footnote{Note that all our results are based on the finishing point of training, which does typically not coincide with the bump.}.

\figmatrix{20 20 20 20}{0.32}{Task Confusion matrices computed on the test set after training of the last task for RAR on Split-MiniImagenet (20 tasks), final training model (Left), RAR+EMA model (Middle), and  RAR+EMA model taken at the tip of the bumps observed in Figure \ref{fig:comparison_imnet} (Right). Note the drop in task-recency bias from RAR (a) to RAR+EMA (b) as a consequence of ensembling.}

\textbf{Comparison with the gains in online i.i.d setting:} When applied on the reference methods $i.i.d$ and $i.i.d_{w/tr}$, the use of EMA model at evaluation also improves the results by a good margin, showing that the gains obtained in continual learning are not only due to continual-learning based improvements (like reducing the task-recency bias), but also on more general online-learning improvements. However, we observe in general higher gains in continual learning than in the i.i.d setting, except on Split-Cifar10, where they are equivalent, showing that the adaptation of the momentum parameter to the distribution drift speed is essential to get continual learning related gains. To illustrate the higher gains in continual learning, we highlight the gap in final accuracy between previous state-of-the-art method (RAR), and the $i.i.d_{w/tr}$ baseline, along with the EMA augmented versions (See Figure~\ref{fig:barplots}). We see that for two of the studied datasets, the gap between $RAR$ and $i.i.d_{w/tr}$ is reduced by the use of the EMA model. For Split-Cifar100, an initial 4.7\% gap is reduced to a 1.8\% gap, while for Split-MiniImagenet, an initial 3.2\% gap is reduced to a 1.5\% gap.

%We observe across all datasets that the gains obtained by combining methods that use input transforms and the EMA model are more important than when not using them (See gains between $i.i.d_{w/tr}$ and $i.i.d_{w/tr}^{+EMA}$ and $RAR$ and $RAR^{+EMA}$)

\textbf{Effect on the stability metrics:} Finally, for all methods and on all datasets, the $AAA$ and the $\operatorname{WC-Acc}$ are greatly improved by the use of EMA, which shows that aside from raising the accuracy, the EMA model offers an important stability boost. We illustrate this effect by showing one single task accuracy curve along with the $\operatorname{WC-Acc}$ curve during training in Figure~\ref{fig:stab_ace} and Figure~\ref{fig:stab_rar}. We see that in both cases both the fluctuations due to small-batch training and the bigger fluctuations due to task shift are reduced by the use of the EMA ensemble. We also present a detailed analysis of stability at the level of a single task shift in Appendix (Figure \ref{fig:analysis_rar}). In general, the biggest increase in $\operatorname{WC-Acc}$ also correspond to the biggest decrease in \textit{Relative Accuracy Gap} ($RAG$), and is significant, confirming that the increase in $\operatorname{WC-Acc}$ is not due to an increase in average accuracy, but due to a better stability.

% Stability gap curves for ER and ER-ACE

% Table with comparison ER, ER-ACE, DER, MIR, GSS?

% Table with results for less memory and Mb couts 10 0000 * 32 * 32 * 3 = 31Mb, 1 model = 4Mb. 10 000 - 8700 mem. For Imnet it's even less (Maybe Appendix)

% Comparison iid gain vs cl gain

\section{Conclusion}
We investigate the effect of employing temporal ensembling methods, such as EMA, in online continual learning. This direction is particularly interesting because of the distinctive nature of this combination. In online continual learning, temporal ensembles offer the potential to combine models from various training tasks, leading to novel dynamics that cannot be achieved in classical offline learning, where every model ensemble is trained on the same distribution.
In the experiments, we show that temporal ensembles can greatly improve continual learning performance and stability. To circumvent the increased memory requirements for the usage of ensembles, we propose to use a memory efficient ensembling solution for online continual learning. We report results using this method in combination with other state-of-the-art methods and conclude that this method consistently increases the final performance and overall stability of several replay methods,  closing in on the performance that can be reached in the $i.i.d$ setting. Most surprisingly, we do this without affecting the training process but just by ensembling models from the training trajectory.

We hope that this work inspires the design of more robust ensembling methods for continual learning. In particular, it would be important to find a similarly efficient method that allows to decorrelate the number of training iteration from the number of tasks, since it could then be applied for arbitrary number of iterations per task while similarly covering as many previous tasks as possible. Another future direction could focus on impacting the training using such kind of ensembles, for instance by combining it with distillation \footnote{We provide comments about the application of distillation in combination with EMA model in the Appendix (Section~\ref{text:mtd}).}.

\textbf{Acknowledgements:} We acknowledge the support of the Grant PID2019-104174GB-I00 funded by MCIN/AEI/ 10.13039/501100011033 and Grant PID2021-128178OB-I00 funded by MCIN/AEI/ 10.13039/501100011033 and by ERDF A way of making Europe, Ramón y Cajal fellowship Grant RYC2019-027020-I funded by MCIN/AEI/ 10.13039/501100011033 and by ERDF A way of making Europe, and the CERCA Programme of Generalitat de Catalunya. Antonio Carta was partially supported by the H2020 project TAILOR (952215) Connectivity Fund. 

\bibliography{collas2023_conference}
\bibliographystyle{collas2023_conference}

\appendix
\section{Appendix}

\subsection{Details about hyperparameter choice for EMA model}

\label{text:hp}

\textbf{Hyperparameter choice.} In all of our experiments, we chose a $\lambda$ of 0.99 for the EMA model, and additionally warm up the EMA model in the first few iterations by setting this parameter to 0.9 in the beginning. This parameter tells us about the horizon of the EMA model. Since we were interested in an ensemble that covers several experiences, we chose an horizon parameter that is high enough so that the EMA model gives non-negligible weights to previous tasks. However, this process has a flaw since it requires to know how many iterations the learner is going to perform on a given task, which is not suppose to happen in continual learning or in online learning. Nevertheless, we believe more practical solutions can be applied to tune this parameter but we did not investigate them. We think the choice of this parameter should depend on the speed at which the input distribution changes, which might be captured by some other mechanism, we leave this direction to future works. In our experiments, we saw that a relatively wide range of parameters between 0.95 and 0.995 were working correctly and always boosting the accuracy in more or less significative ways. We present results for varying $\lambda$ on Split-Cifar100 in Figure~\ref{fig:comparison_lambdas}. We see in that figure that increasing $\lambda$ until the accuracy bump discussed in Section~\ref{text:results} disappears is overall beneficial for the final average accuracy, but this might not be possible to do in practice when not knowing the amount of data to be received from each class. Also, higher $\lambda$ values, while beneficial for the final average accuracy, affect negatively the accuracy for earlier tasks. One work around to that problem would be to tune $\lambda$ progressively according to the number of already seen classes vs the number of new coming classes.

\textbf{The accuracy bumps} that we observe in various figures when looking at the average accuracy indicate that the gains that we report in the tables could be bigger if we would retain the model at the tip of the bump. However, this requires the use of a hold-out validation memory and an additional copy of the model on disk. In Figure~\ref{fig:confmatrix}, we retain the model using the above technique of validation memory and observe that these bumps correspond to a lower task-recency bias in the saved EMA model.

\textbf{Trade-off.} This choice of lambda also highlights a problem of this method that makes it difficult to export to non-online learning. It is its dependency on the number of training iterations. Indeed, choosing a higher $\lambda$ increases the time horizon of the method, but also has negative effects for several reasons. First of all, it is not clear how high of a $\lambda$ can be chosen before the performance of the ensemble decreases since giving too much weights to single models that are too far in the weight space from the last model might break the assumption that summing models in the weight space leads to a model with accuracy equivalent to the ensemble of these two models \cite{subspaces}. This assumption is believed to be true as long as two models are connected by a linear path of low loss, and it is possible that two models that are far away from each other do not respect this constraint. Secondly, it would be a nice addition to be able to sum models obtained after each training batch, or even every training task (instead of each training iteration, since several iterations can be performed on the same training batch in online learning). However for the same reason invoking linear connectivity properties, it is not clear that this would work any better than increasing the momentum value in the computation of the exponential moving average.

\textbf{Task Covering.} For the chosen parameter value, on Split-Cifar100, we can compute the total amount of weight assigned to each task. From Eq.~\ref{eq:explicit}, we deduce the weight for a single member in the ensemble $\theta^{i}$ is $(1-\lambda)\lambda^{t-i}$ where $t$ is the iteration of the current training model. We can sum this over the iterations covering one task. If we place ourselves at the end of one task, with $\lambda = 0.99$  (value used in our experiments) we get $88\%$ of the weights in the last task, $9\%$ of the weight in the previous task, and $1\%$ in the second last task. This indicates that a majority of the weights are assigned to later tasks (in particular from last task to second last task). However, we argue here that as the mean can be highly influenced by outliers, the exponential moving average is nothing else than a weighted mean, and can be influenced by outliers as well. Since the distance in the weight space between models from different tasks is more important than the one between models from the same task, it is possible that earlier models influence more strongly the current moving average than what we could believe by looking at the weights value.

\figlambdas{10 10 10 10}{1.0}{Comparison of the results obtained for various $\lambda$ values of the EMA model on Split-Cifar100 (20 tasks) for ER. Each curve depicts the average accuracy of the EMA model using a different $\lambda$ on all tasks seen so far. The value we chose in our experiments (0.99) is not optimal but nor do we have a way of chosing the optimal one. Mean and standard deviation are computed over 3 seeds.}

\subsection{Details about the motivational experiment}

\label{text:explained_motiv}

\figema{10 10 0 10}{0.48}{Schema of the motivational experiment ensemble (Left) and of the Exponential Moving Average ensemble (Right). For the EMA ensemble, a continuum of models is inserted in the ensemble which only occupies as much space as one additional model in the memory. Weights of previous model checkpoints decrease exponentially. This procedure permits to cover a spectrum of different tasks in online continual learning.}

Figure~\ref{fig:motivation} shows the comparison between ensembles of models that cover a different number of tasks. To perform this comparison, we first train a model continually using a dedicated continual learning method (ER and ER-ACE in that case), along the way, we save model checkpoints every 10 iterations, leaving about 460 model checkpoints at the end of learning the task (22 models per task in the case of Split-Cifar100 20 tasks). Then, for each number of tasks covered, we select a subset of models that cover no more than that number of tasks, and sample 20 models from that subset that we join into an ensemble for which we record the test performance. We sample 10 of such ensembles for every x-axis value to get mean and confidence interval that we report in the figure. We always add a model in the ensemble which is the last training checkpoint, so that we can always give an accuracy number to later classes when sampling models. The ensembling of models coming from different tasks is done in the following way. We first compute the output probabilities that each model gives to the input. We then compute the per-class mean probability, it is possible that one class is predicted different number of times than another class by the ensemble, so we take that into account and divide the sum of the probabilities for that class by the number of models in the ensemble that can predict that class.

\subsection{Supplementary material about the stability gap}

Here we show several continual evaluation curves (see Figure \ref{fig:comparison_cifar100}). All of these curves were created by evaluating the model on a hold-out validation set which is $5\%$ of the training set size. The evaluation is performed after training for 3 iterations on each mini-batch before dropping it, as required by the constraints of online learning. 

\begin{figure}
    \centering
    \includegraphics[width=1.0\textwidth]{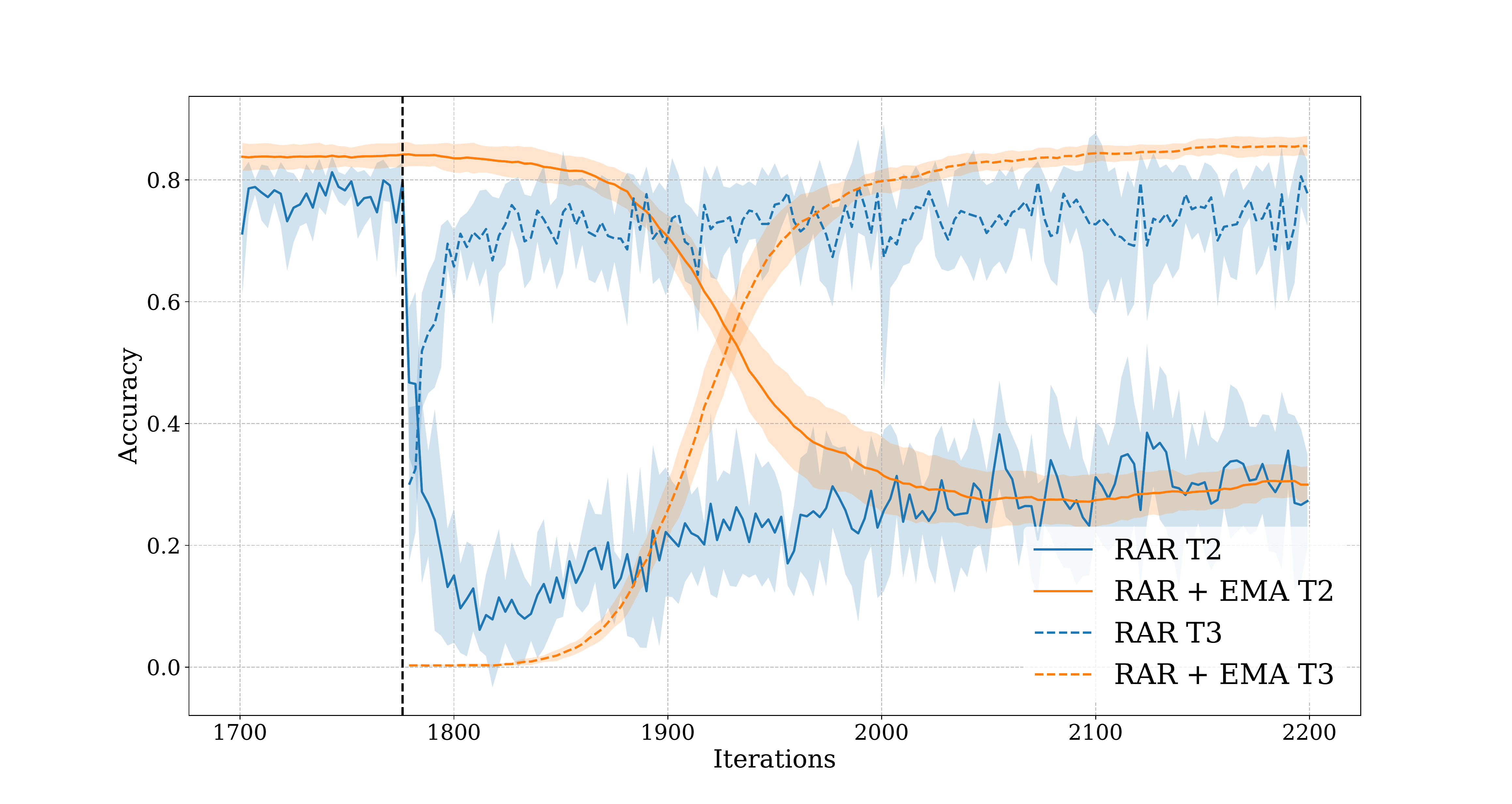}
    \caption{Split-Cifar10, validation accuracy on task 2 data and task 3 data at task shift (indicated with dotted black line). In blue, the accuracy for the training model, in orange, the accuracy for the EMA model. Solid lines indicate accuracy on task 2 data while dotted lines indicate accuracy on task 3 data.}
    \label{fig:analysis_rar}
\end{figure}

In Figure \ref{fig:analysis_rar}, we compare the accuracy of RAR \citep{rar} and RAR+EMA on the data of two subsequent tasks during the task shift. We see that the current training model instantly loses accuracy on previous task at the task shift before regaining part of this accuracy later in training. This results in an important \textit{stability gap} since the accuracy on previous task reaches a low point during training before going back to "normal". This observation is coherent with the one made in \cite{lange2023continual}. The orange curves indicate the performance of the EMA model on the same tasks. We see that the EMA models takes longer to get good performance on new task but ends up getting better performance than the training model. EMA models also displays improved stability (performance on previous task is smoothly going from initial to final performance).

\figcifarcien{10 10 10 10}{1.0}{Comparison of the average accuracy on the validation data of three methods trained on Split-Cifar100 (20 tasks), and their EMA augmented version. EMA performance is indicated with the dotted lines. We report the Mean and standard deviation over 6 runs}

% Comparison, single experience accuracies on Split-Cifar100, ER no EMA, ER with EMA

% Same with another method

% Same with another dataset

\subsection{Attempts at using EMA model for distillation}

\label{text:mtd}

\cite{tarvainen2017mean} initially introduced the EMA model in order to use it as a teacher model in semi-supervised learning. They find that distilling knowledge from the EMA model to the student is beneficial in that setting. We likewise tried to apply distillation by using the EMA model as a teacher, however we found that gains obtained by this method (when present) were not robust enough to be reported. In Table~\ref{tab:mtd}, we report the student and teacher (EMA) performance when using, and not using, Mean-Teacher distillation (See Eq.~\ref{eq:mtd}). We found that some improvements could be observed in combination with simple ER on Split-Cifar100 both in terms of student and teacher performance, however these improvements did not generalize to Split-MiniImagenet, neither do they combine well with a stronger method like RAR. Notably, in the case of $ER_{+MTD}$ for Split-MiniImagenet and $RAR_{+MTD}$ for both datasets, we observe that the final accuracy of the student is only slightly modified by the distllation process (sometimes increased, sometimes decreased), but the accuracy of the teacher is decreased consequently to its application, this indicates that the distillation process might reduce the diversity of the EMA ensemble by pulling models from the training trajectory closer from one another.

\begin{equation}
    \label{eq:mtd}
    L_{CE}(f_{\theta}(x), y) + \alpha L_{CE}(f_{\theta}(x), f_{\theta_{EMA}(x)})
\end{equation}

\begin{table}[h!]
    \centering
    \begin{tabular}{c|c|c}
         \hline
         Method & Split Cifar100 &  Split MiniImagenet \\
          & Acc &  Acc \\
         \hline
         $ER$ &  9.9 $\pm$ 0.6 & 26.2 $\pm$ 0.2 \\
         $ER$ (EMA) & 14.0 $\pm$ 0.5 & 36.3 $\pm$ 1.1 \\
         \hline
         $ER_{+MTD}$ &  14.7 $\pm$ 0.5 & 27.1 $\pm$ 2.1 \\
         $ER_{+MTD}$ (EMA) & 19.0 $\pm$ 0.4 & 33.5 $\pm$ 1.4 \\
         \hline
         $RAR$ & 27.6 $\pm$ 1.3  & 29.1 $\pm$ 0.8 \\
         $RAR$ (EMA) & 35.4 $\pm$ 1.2 & 38.4 $\pm$ 0.8 \\
         \hline
         $RAR_{+MTD}$ & 27.0 $\pm$ 1.0 & 27.1 $\pm$ 2.5 \\
         $RAR_{+MTD}$ (EMA) & 32.7 $\pm$ 0.9 & 33.4 $\pm$ 2.5 \\
         \hline
    \end{tabular}
    \caption{Mean Teacher distillation results for teacher and non teacher on Split-Cifar100 (20 Tasks) (Left) and Split-MiniImagenet (Right). Mean and standard deviation are reported over 6 seeds.}
    \label{tab:mtd}
\end{table}

\section{Comparison of memory usage with and without the use of EMA model}

We include a comparison of memory overhead induced by the use of EMA across all datasets (see Table~\ref{tab:memory}). Using EMA model at evaluation requires to store an additional model. In terms of relative memory usage increase, it is more interesting to use EMA model for larger datasets for which exemplars require more memory to store (MiniImagenet) than for small datasets like Cifar. The methods we compare in this article do not differ significantly from each other in terms of memory usage since they all use replay with the same number of exemplars, this is why we omit them from the comparison.

\begin{table}[h!]
    \centering
    \begin{tabular}{c|c|c|c|c|c|c}
         \hline
         Method & \multicolumn{2}{c|}{Split Cifar10} & \multicolumn{2}{c|}{Split Cifar100} &  \multicolumn{2}{c}{Split MiniImagenet} \\
         Memory & Exemplars & Model & Exemplars & Model & Exemplars & Model \\
         \hline
         w/o EMA & 3 Mb & 4 Mb & 6 Mb & 4 Mb & 211 Mb & 4 Mb \\
         w/ EMA & 3 Mb & 8 Mb & 6 Mb & 8 Mb & 211 Mb & 8 Mb \\
    \end{tabular}
    \caption{Memory usage for the considered datasets and for the exemplars used in our experiments (1000, 2000 and 10000 for the three datasets respectively).}
    \label{tab:memory}
\end{table}

\begin{figure}
    \centering
    \includegraphics[width=1.0\textwidth]{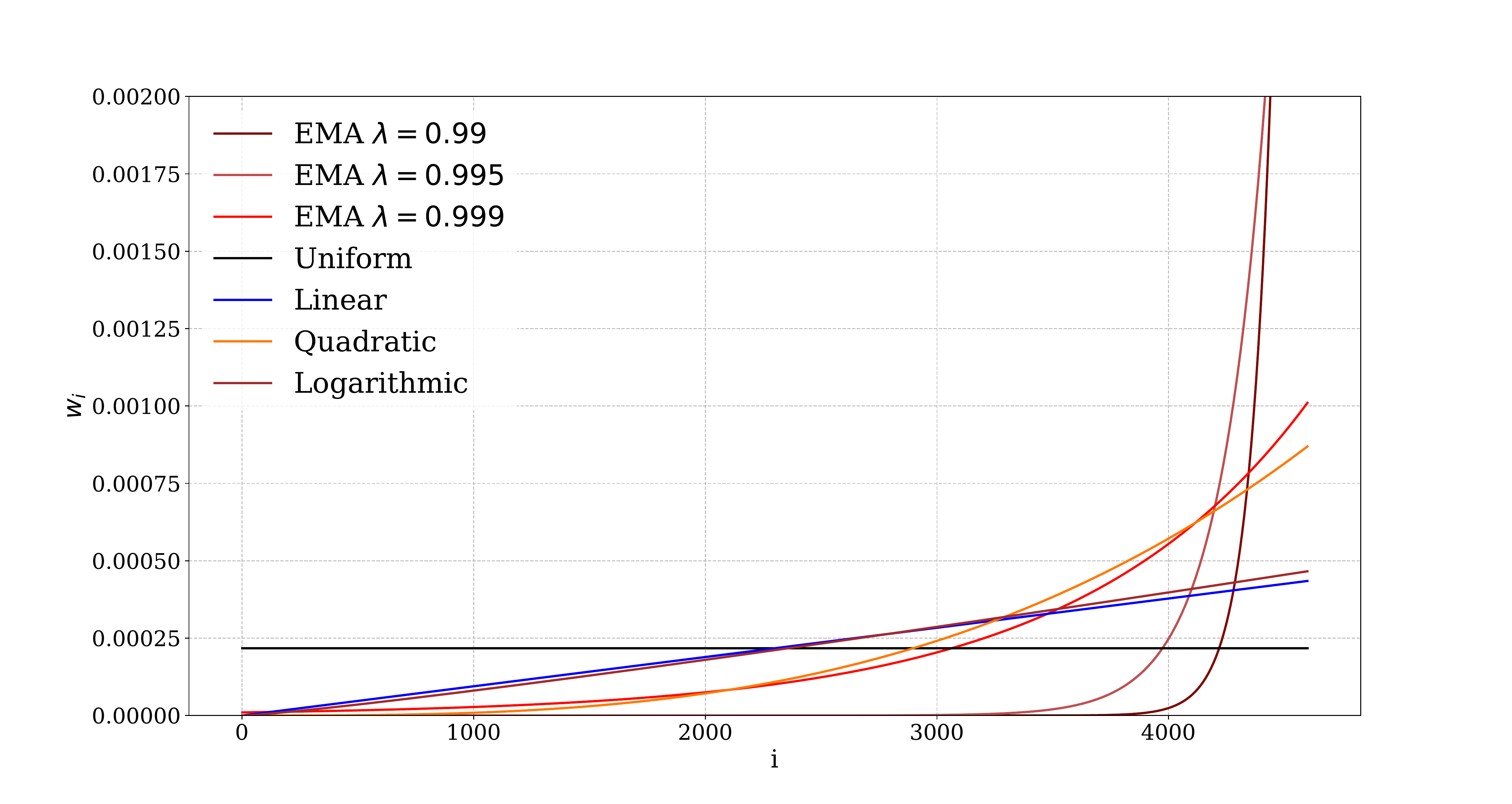}
    \caption{Curve showing the evolution of different weighting schemes ($w_i$) under the form described in Equation~\ref{eq:explicit_other}. We compare the performance of these weighting schemes in Section \ref{text:method}. We observe here that EMA with a high lambda leads to a weighting that looks similar to quadratic weighting. To draw these curves we place ourselves at the last training iteration ($\theta_{ensemble}^{t_{last}}$)}
    \label{fig:weighting_schemes}
\end{figure}

\begin{figure}
    \centering
    \includegraphics[width=1.0\textwidth]{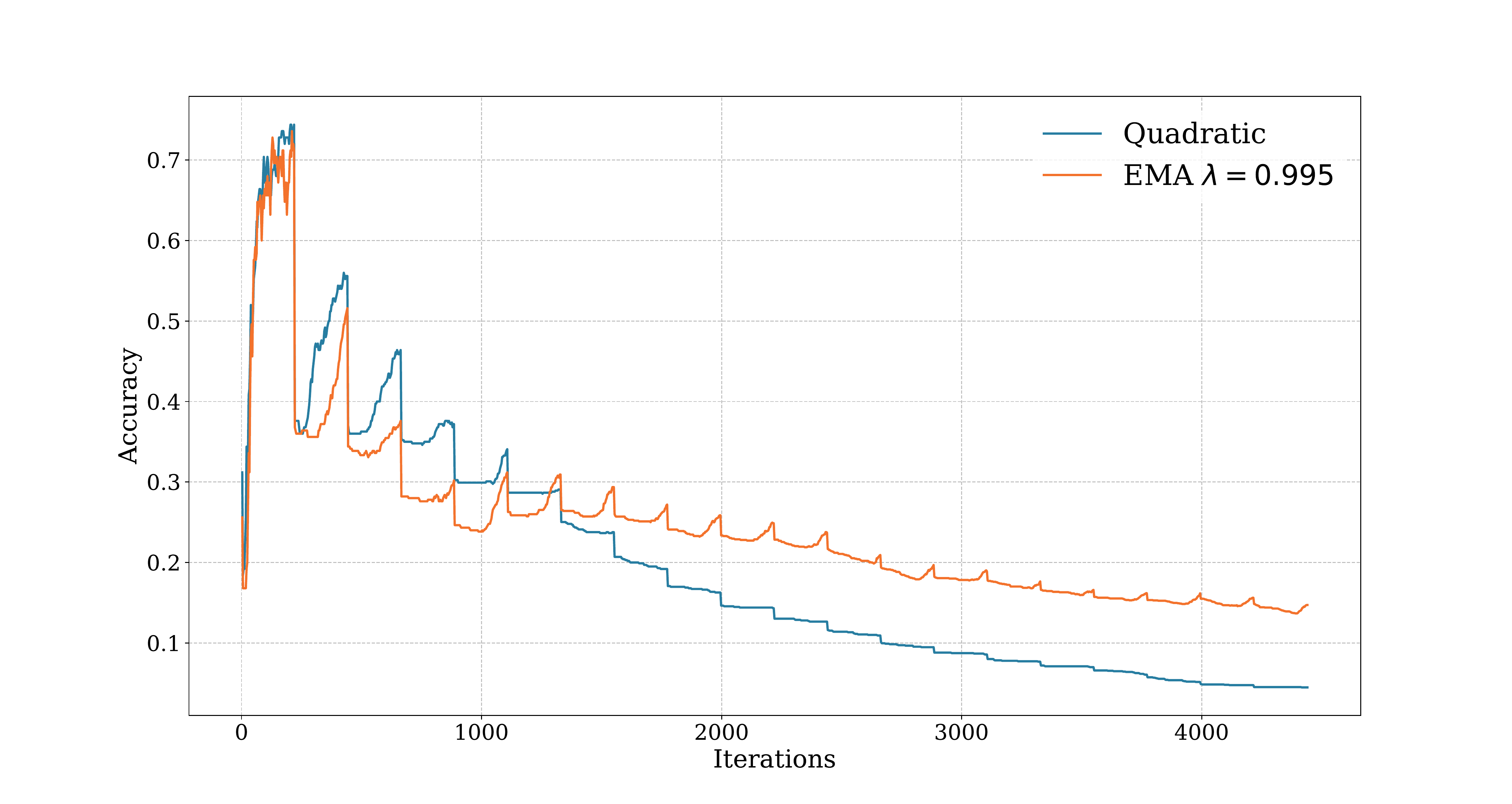}
    \caption{$\operatorname{WC-ACC}_t$ for Quadratic vs EMA with $\lambda = 0.995$ in combination with ER on Split-Cifar100 dataset. We see that the quadratic weighting scheme gives more interesting stability (in terms of $\operatorname{WC-ACC}_t$) in the first few tasks, but then fails short compared to the EMA ensemble.}
    \label{fig:wc_acc_quad}
\end{figure}

\end{document}